\documentclass{article}

\usepackage[preprint]{neurips_2022}

\usepackage[utf8]{inputenc} % allow utf-8 input
\usepackage[T1]{fontenc}    % use 8-bit T1 fonts
\usepackage{hyperref}       % hyperlinks
\usepackage{url}            % simple URL typesetting
\usepackage{booktabs}       % professional-quality tables
\usepackage{amsfonts}       % blackboard math symbols
\usepackage{nicefrac}       % compact symbols for 1/2, etc.
\usepackage{microtype}      % microtypography
\usepackage{xcolor}         % colors

\usepackage{amsmath,amsfonts,bm}

\providecommand{\ie}{\textit{i.e.}}

\def\eqref#1{equation~(\ref{#1})}
\def\1{\bm{1}}

\def\vc{{\bm{c}}}

\def\vf{{\bm{f}}}

\def\vx{{\bm{x}}}

\def\vz{{\bm{z}}}

\def\mI{{\bm{I}}}

\DeclareMathAlphabet{\mathsfit}{\encodingdefault}{\sfdefault}{m}{sl}
\SetMathAlphabet{\mathsfit}{bold}{\encodingdefault}{\sfdefault}{bx}{n}

\def\gN{{\mathcal{N}}}

\newcommand{\E}{\mathbb{E}}

\usepackage{tabularx}
\usepackage{graphicx}
\usepackage{subfigure}
\usepackage{multirow}
\usepackage{makecell}
\usepackage{gensymb}
\usepackage{pifont}
\usepackage{pict2e}
\usepackage{titletoc}
\usepackage{amsmath,amssymb,amsthm}
\title{Accelerating Diffusion Models via Early Stop of the Diffusion Process}

\newcommand{\AuthorSpace}{\hspace{1.2em}}
\author{%
 Zhaoyang Lyu$^{1,2}$\AuthorSpace{} Xudong Xu$^{1,2}$\AuthorSpace{} Ceyuan Yang$^{1,2}$\AuthorSpace{} Dahua Lin$^{1,2}$\AuthorSpace{} Bo Dai$^{2}$ \\
 $^1$The Chinese University of Hong Kong \\
 $^2$Shanghai AI Laboratory \\
 \texttt{lyuzhaoyang@link.cuhk.edu.hk, xx018@ie.cuhk.edu.hk} \\
\texttt{yc019@ie.cuhk.edu.hk, dhlin@ie.cuhk.edu.hk, daibo@pjlab.org.cn}
}

\begin{document}

\maketitle

\begin{abstract}
Denoising Diffusion Probabilistic Models (DDPMs) have achieved impressive performance on various generation tasks.
By modeling the reverse process of gradually diffusing the data distribution into a Gaussian distribution, generating a sample in DDPMs can be regarded as iteratively denoising a randomly sampled Gaussian noise.
However, in practice DDPMs often need hundreds or even thousands of denoising steps to obtain a high-quality sample from the Gaussian noise, leading to extremely low inference efficiency.
%Denoising Diffusion Probabilistic Models (DDPMs) have achieved impressive performance on image generation.
%However, their sampling process is extremely slow, as they need hundreds or even thousands of denoising steps to generate a clean image.
%DDPMs define a diffusion process that gradually adds noise to a clean image until it turns into a Gaussian noise.
%To sample clean images, DDPMs use a neural network to simulate the reverse process of the diffusion process. 
%The network iteratively denoises a Gaussian noise to a clean image.
% Prior works achieve successful acceleration of DDPMs by sampling on a subsequence of the original denoising process, but the sample quality drops significantly with a small number of denoising steps.
In this work, we propose a principled acceleration strategy, referred to as Early-Stopped DDPM (ES-DDPM), for DDPMs.
The key idea is to stop the diffusion process early
where only the few initial diffusing steps are considered and the reverse denoising process starts from a non-Gaussian distribution.
By further adopting a powerful pre-trained generative model, such as GAN and VAE, in ES-DDPM,
sampling from the target non-Gaussian distribution can be efficiently achieved by diffusing samples obtained from the pre-trained generative model.
%In this work, we propose to adopt early stop in the diffusion process of a DDPM to obtain a noisy image instead of a Gaussian noise.
%Correspondingly, in the sampling process, we first sample a noisy image using a fast generative model like VAE or GAN, then use the DDPM to denoise the noisy image to a clean image in fewer steps than the original DDPM.
In this way, the number of required denoising steps is significantly reduced.
In the meantime, the sample quality of ES-DDPM also improves substantially,
outperforming both the vanilla DDPM and the adopted pre-trained generative model.
On extensive experiments across CIFAR-10, CelebA, ImageNet, LSUN-Bedroom and LSUN-Cat,
ES-DDPM obtains promising acceleration effect and performance improvement over representative baseline methods.
Moreover,
ES-DDPM also demonstrates several attractive properties, 
including being orthogonal to existing acceleration methods,
as well as simultaneously enabling both global semantic and local pixel-level control in image generation.
%Surprisingly, we find the combined model can even outperform the original DDPM in terms of sample quality. 
%We also demonstrate that our method can be easily coupled with other DDPM acceleration methods to achieve further acceleration.
%We conduct extensive experiments on CIFAR-10, CelebA, ImageNet, LSUN-Bedroom and LSUN-Cat datasets to verify the effectiveness of our acceleration method, and achieve the best acceleration effect in many cases.
%We also find that our method not only accelerates the sampling process of a DDPM, but also the training process.
%Another benefit of our method is that the combined model can control both high-level semantics and local details of the generated image.

\end{abstract}
\section{Introduction}
% Denoising Diffusion Probabilistic Models (DDPMs) are a class of generative models that have recently achieved impressive generation results in multiple domains.
% It is first introduced by \citet{sohl2015deep}, and then improved by \citet{ho2020denoising}. 

Denoising Diffusion Probabilistic Models (DDPMs)~\citet{sohl2015deep} are a class of generative models that have received growing attention in recent years,
due to their promising results in both unconditional and conditional generation tasks, such as image generation~\citep{ho2020denoising, dhariwal2021diffusion, nichol2021glide}, image manipulation and restoration \citep{nichol2021glide, meng2021sdedit, saharia2021image, saharia2021palette}, audio generation~\citep{chen2020wavegrad, kong2020diffwave}, as well as 3D shape generation and completion~\citep{luo2021diffusion, zhou20213d, lyu2021conditional, zhou20213d}.
DDPMs regard the generation procedure as the reverse of a diffusion process, which gradually adds noises to data samples and transforms the data distribution into a Gaussian distribution.
Therefore,
synthesizing a sample from DDPMs is achieved by denoising a randomly sampled Gaussian noise iteratively.
%They also demonstrate promising results in many conditional generation tasks such as image editing~\citep{}, image super-resolution~\citep{}, image restoration~\citep{}, audio upsampling~\citep{lee2021nu} and point cloud completion~\citep{lyu2021conditional, zhou20213d}. 
%DDPMs define a diffusion process that gradually adds noise to a clean image until turning it into a Gaussian noise.
%Then DDPMs use a neural network to approximate the reverse process of the diffusion process. 
%To sample images from a DDPM, we use the network to denoise a Gaussian noise iteratively until turning it into a clean image.
% \textcolor{blue}{briefly introduce DDPM diffusion and generation process.}

While DDPMs show great potential in various generation tasks,
one major issue of them is their intrinsic low inference efficiency,
as obtaining a high quality sample from the vanilla DDPM usually requires thousands of iterative denoising steps, each of which involves a forward evaluation of the underlying neural network.
%However, the sampling process of a DDPM requires hundreds and even thousands of denoising steps in the form of forward evaluations of a Unet-like neural network, which is extremely expensive.
To accelerate DDPMs,
previous attempts~\citep{ho2020denoising, kong2021fast, nichol2021improved} propose to integrate multiple standard denoising steps into a single jumping denoising step.
Although the number of denoising steps is reduced, 
such integrated denoising steps break the Gaussian assumption of DDPMs' denoising process~\citep{xiao2021tackling, sohl2015deep},
resulting in dropped sample quality.
%only a subsequence of the original denoising steps by skipping  
%have been proposed to accelerate DDPMs without retraining the network. 
%They typically generate sample with only a subsequence of the original denoising steps by making jumping steps in the reverse steps, but sample quality inevitably drops as the number of denoising steps decrease.
%This is because reducing the number of denoising steps breaks the Gasussian assumption of the denoising distribution in the reverse process of a DDPM~\citep{xiao2021tackling, sohl2015deep}. 
% \citet{salimans2022progressive} propose to progressively distill a trained DDPM for fast sampling.
On the other hand,
acceleration of DDPMs is also indirectly achieved by \citep{preechakul2021diffusion, pandey2022diffusevae},
which inject image features or VAE-generated images into the denoising process as additional conditions.
Such conditioned denoising processes are more effective than the original one, requiring fewer denoising steps to obtain a high-quality sample.
However, their acceleration effect is often limited in practice since there are no modifications to the vanilla conditional DDPM generation framework in these approaches.
It thus remains a challenging task to accelerate DDPMs in a principled way without compromising the quality of synthesized samples,
especially for unconditional generation or class-conditioned generation,
where no strong conditioning signal is available. 
%It remains a challenging task to obtain high quality samples with a small number of denoising steps, especially when conditioning signal is non-existent or weak, \eg, unconditional image generation or class-conditional image generation.

\begin{figure}[t]
    \centering
    \vspace{-4.8em}
    % \label{fig:mlp}
    \includegraphics[width=0.95\textwidth]{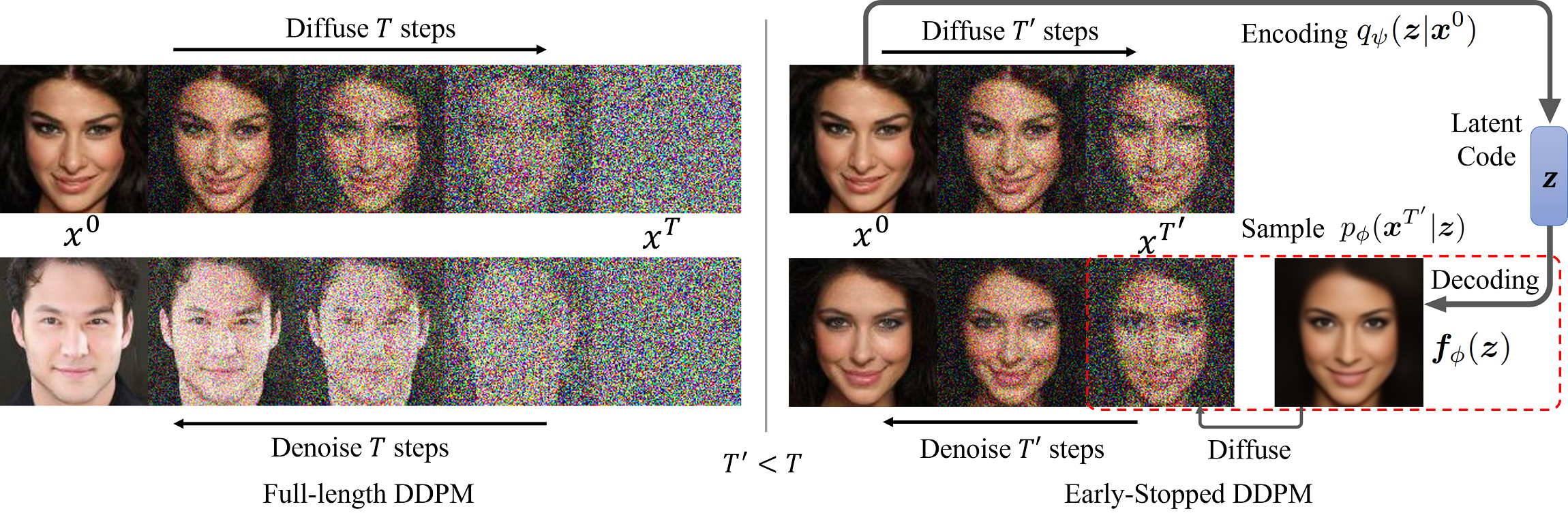}
    \vspace{-1.2em}
    \caption{Compare Early-Stopped DDPM (ES-DDPM) with the full-length DDPM.}
    \label{fig:es_ddpm_vs_ddpm}
    \vspace{-1em}
\end{figure}

In this work, we propose a novel acceleration strategy for DDPMs that not only boosts the efficiency of DDPMs without breaking any of the assumptions,
but also, as an additional benefit, improves the generation quality of DDPMs.
The core idea of our strategy, dubbed as \textbf{Early-Stopped DDPM (ES-DDPM)}, is stopping the diffusion process early as shown in Figure~\ref{fig:es_ddpm_vs_ddpm}.
Instead of diffusing the data distribution into a Gaussian distribution via hundreds to thousands of iterative steps,
ES-DDPM considers only the initial few diffusion steps,
so that the reverse denoising process starts from a non-Gaussian distribution $q$,
leading to a significantly reduced number of denoising steps.
One remaining issue is ES-DDPM requires sampling from the non-Gaussian distribution $q$ to generate data samples,
yet $q$ has no closed form formulations.
To overcome this issue,
we observe that $q$ is diffused from the real data distribution,
in the meantime many of existing generative models provide a good approximation of the real data distribution.
Consequently,
we further equip ES-DDPM with a pre-trained generative model (e.g. GAN~\citep{karras2019style} and VAE~\citep{kingma2013auto}) 
so that sampling from $q$ becomes diffusing samples obtained from the pre-trained generative model.
Since both the sampling and the diffusion operation can be easily achieved in a single step,
the pre-trained generative model brings only minor computational overhead to ES-DDPM.

The proposed ES-DDPM bears several important advantages.
\textbf{1)} The number of denoising steps in ES-DDPM is significantly reduced (from 1000 steps to 100 steps in the best case of our experiments). 
% it demonstrates much stronger acceleration effect than previous approaches.
Moreover,
such acceleration effect is achieved without sacrificing the quality of generated samples. 
In fact, by combining a pre-trained generative model, 
ES-DDPM outperforms both the vanilla DDPM as well as the pre-trained generative model in terms of sample quality.
\textbf{2)} {Unlike the acceleration strategies based on jumping denoising steps that only accelerates the generation process during inference,} ES-DDPM provides a principled way to accelerate DDPMs in both training and inference.
\textbf{3)} While major properties of the vanilla DDPM is preserved in ES-DDPM, it is orthogonal to most existing acceleration strategies of DDPMs. As shown in our experiments, ES-DDPM coupled with other acceleration strategies results in further improved acceleration effect.
\textbf{4)} ES-DDPM not only maintains the properties of the vanilla DDPM, it also enjoys the properties of the involved pre-trained generative model. For instance, when combined with a StyleGAN2 for unconditional image generation, ES-DDPM enables both global semantic control and local pixel-level control over generated images, as shown in our experiments.
%We find that ourvolvement method bears the following advantages:
%\textbf{1)} We can accelerate the generation process of a DDPM without a drop in sample quality. In fact, we find the combined generative model can outperform both the original VAE or GAN and the DDPM in terms of sample quality, while being faster than the original DDPM.
%\textbf{2)} Our method not only accelerates the sampling process of a DDPM. but also accelerate the training process of the DDPM.
%\textbf{3)} Our method is orthogonal to most previous DDPM acceleration methods. It can be easily coupled with other acceleration methods to further accelerate the sampling process of a DDPM.
%\textbf{4)} Our method combines both VAE or GAN's advantage and DDPM's advantage in terms of controllable generation. It can semantically control the content of the generated image, while at same time allow pixel-level control of the generated image.

\section{Background on DDPMs}
In Denoising Diffusion Probabilistic Models (DDPMs), the diffusion process is defined as
\begin{equation}
\label{eqn:diffusion_process}
    q(\vx^{1:T}|\vx^0) = \prod_{t=1}^T q(\vx^t|\vx^{t-1}),
    \text{ where }
    q(\vx^t|\vx^{t-1})=\gN(\vx^t;\sqrt{1-\beta_t}\vx^{t-1},\beta_t \mI),
\end{equation}
where $\beta_t$'s are some predefined small constants, and $\vx^{1:T}$ denotes the set of variables $\{\vx^1, \vx^2,\cdots, \vx^T\}$. Similar notations are used to denote a set of variables in rest part of the paper.
In the diffusion process, we gradually add noise to the clean image $\vx^0$. $T$ is set to be a sufficiently large number so that $\vx^T$ is close to a Gaussian noise. $T=1000\sim4000$ is a typical choice for most works.
The reverse process is defined as 
\begin{align}
\label{eqn:reverse_process}
    p_{\bm{\theta}}(\vx^{0:(T-1)}|\vx^T)=\prod_{t=1}^T p_{\bm{\theta}}(\vx^{t-1}|\vx^t),
    \text{ where }
    p_{\bm{\theta}}(\vx^{t-1}|\vx^t) = \gN(\vx^{t-1};\bm{\mu}_{\bm{\theta}}(\vx^t, t), \sigma_t^2\mI).
\end{align}
The variance $\sigma_t^2$'s can be set as time-step dependent constants or learned by a neural network.
The mean is parameterized by a neural network $\bm{\mu}_{\bm{\theta}}(\vx^t, t)$.
The neural network is trained to simulate the reverse process of the diffusion process defined in Equation~\ref{eqn:diffusion_process}.
To generate an image from the reverse process, we first sample $\vx^T$ from the Gaussian distribution, and then sample $\vx^{t-1}$ from $p_{\bm{\theta}}(\vx^{t-1}|\vx^t)$ for $t=T,T-1,\cdots,1$. $\vx^0$ is the image that the DDPM generates.

The generation process of a DDPM is extremely slow as it need to sample from the transition distribution $p_{\bm{\theta}}(\vx^{t-1}|\vx^t)$ iteratively, which involves hundreds or even thousands of evaluations of the output of the neural network $\bm{\mu}_{\bm{\theta}}(\vx^t, t)$.

\section{Methodology}
It is necessary to accelerate the sampling process of the DDPM for any practical usage.
Our key insight is that we do not need to diffuse a clean image thousands of steps to turn it into a Gaussian noise as shown in Equation~\ref{eqn:diffusion_process}. 
We can adopt early stop in the diffusion process as shown in Figure~\ref{fig:es_ddpm_vs_ddpm}. 
We cut the diffusion process in the middle at $t=T'<T$, and use an encoder to encode the noisy image $x^{T'}$ into a low dimension vector $\vz$ in one step.
Correspondingly, in the sampling process, we first sample a latent code $\vz$ from the Gaussian distribution, then use a decoder to sample $\vx^{T'}$ from  $\vz$. Finally, we use the DDPM to denoise $\vx^{T'}$ to a clean image $\vx^{0}$ in fewer steps than the original sampling process.
We name the cut DDPM \textbf{early-stopped DDPM} (\textbf{ES-DDPM}).
% This acceleration strategy is applicable for both unconditional and conditional generation tasks. We elaborate on these two settings in Section~\ref{sec:unconditional_generation} and Section~\ref{sec:conditional_generation}, respectively. 

% \subsection{Unconditional Generation}
% \label{sec:unconditional_generation}
Our encoding process, which encodes the clean image $\vx^0$ to a low dimension latent code $\vz$, is defined as
\begin{align}
\label{eqn:encoding_process}
    q(\vx^{1:T'}, \vz | \vx^0) = q(\vx^{1:T'} | \vx^0) q_{\psi}(\vz | \vx^{1:T'}, \vx^0) = q(\vx^{1:T'} | \vx^0) q_{\psi}(\vz | \vx^0),
\end{align}
where $q_{\psi}(\vz | \vx^0)$ is the encoder. We set it to a Gaussian distribution whose mean and standard deviation are parameterized by a neural network.
$q(\vx^{1:T'} | \vx^0)$ is a partial diffusion process of Equation~\ref{eqn:diffusion_process}, and it is defined as
\begin{align}
\label{eqn:partial_diffusion_process}
    q(\vx^{1:T'} | \vx^0) = \prod_{t=1}^{T'} q(\vx^t|\vx^{t-1}),
    \text{ where }
    q(\vx^t|\vx^{t-1})=\gN(\vx^t;\sqrt{1-\beta_t}\vx^{t-1},\beta_t \mI).
\end{align}
In this partial diffusion process, we do not need $T'$ to be large enough to guarantee that $x^{T'}$ is close to a Gaussian noise. 
% Instead, we have the encoder to encodes $x^{T'}$ to a latent code $\vz$, which we assume to follow the Gaussian distribution.

We define the sampling process as
\begin{align}
\label{eqn:generation_process}
    p(\vx^{0:T'}, \vz) = p(\vz) p_{\phi}(\vx^{T'} | \vz) p_{\theta}(\vx^{0:(T'-1)}|\vx^{T'}),
\end{align}
where $p(\vz)$ is assumed to follow standard Gaussian distribution, $p_{\phi}(\vx^{T'} | \vz)$ is the decoder parameterized by a neural network, and $p_{\theta}(\vx^{0:(T'-1)}|\vx^{T'})$ is the reverse process of the partial diffusion process in Equation~\ref{eqn:partial_diffusion_process}. It is defined as
\begin{align}
\label{eqn:partial_reverse_process}
    p_{{\theta}}(\vx^{0:(T'-1)}|\vx^{T'})=\prod_{t=1}^{T'} p_{{\theta}}(\vx^{t-1}|\vx^t),
    \text{ where }
    p_{{\theta}}(\vx^{t-1}|\vx^t) = \gN(\vx^{t-1};\bm{\mu}_{{\theta}}(\vx^t, t), \sigma_t^2\mI).
\end{align}

Through variational inference, we can prove that
\begin{align}
\label{eqn:ddpm_elbo}
    \log p(\vx^{0}) \geq - (L_{\text{VAE}} + L_{\text{DDPM}}), 
\end{align}
\begin{align}
\text{where }
    L_{\text{VAE}} &= D_{\text{KL}}(q_{\psi}(\vz | \vx^0)||p(\vz)) + \E_{q_{\psi}(\vz | \vx^0)} D_{\text{KL}} (q(\vx^{T'}|\vx^0) || p_{\phi}(\vx^{T'} | \vz)), \label{eqn:l_vae}\\
    L_{\text{DDPM}} &= \sum_{t=2}^{T'} 
    \E_{ q(\vx^t|\vx^0) } 
    D_{\text{KL}}( q(\vx^{t-1}|\vx^{t}, \vx^0) || p_{\theta}(\vx^{t-1}|\vx^t) ) 
    - \E_{q(\vx^{1} | \vx^0)} \log p_{\theta}(\vx^{0}|\vx^1). \label{eqn:l_ddpm}
\end{align}
$D_{\text{KL}}$ denotes Kullback–Leibler divergence. See complete proof in Appendix Section~\ref{sec:elbo_proof}.
To maximize the data log-likelihood $\log p(\vx^{0})$, we need to minimize the two loss terms: $L_{\text{DDPM}}$ and $L_{\text{VAE}}$. 
We will demonstrate that this means we can train an ES-DDPM and a variational autoencoder (VAE) separately, and then combine them together to obtain a new generative model. 
First, The loss term $L_{\text{DDPM}}$ is the same of a normal DDPM, except that we do not need to train the full Markov chain range from $1$ to $T$. 
Instead, we only need to train the first several steps range from $1$ to $T'$. 
A normal DDPM trained on the loss of full $T$ terms would also fulfil the requirements here, but as we will show in later experiments, the training process of an ES-DDPM is faster than a normal full-length DDPM.

Next, we show that the loss term $L_{\text{VAE}}$ amounts to train a VAE on the clean images $\vx^0$ in the dataset.
The first term in $L_{\text{VAE}}$ is the same as in a standard VAE, which requires the encoder to encode clean images to a latent space that follows the Gaussian distribution.
The second term in $L_{\text{VAE}}$ means that we want $p_{\phi}(\vx^{T'} | \vz)$ to match $q(\vx^{T'}|\vx^0)$ as close as possible.
It is proven by previous works~\citep{sohl2015deep, ho2020denoising} that
% $q(\vx^{T'}|\vx^0) \sim N(\vx^{T'}; \sqrt{\bar{\alpha}_t} \vx^0, (1-\bar{\alpha}_t)\mI)$.
\begin{align}
    q(\vx^{T'}|\vx^0) \sim N(\vx^{T'}; \sqrt{\bar{\alpha}_t} \vx^0, (1-\bar{\alpha}_t)\mI), \text{ where } \alpha_t = 1 - \beta_t, \bar{\alpha}_t = \prod_{i=1}^t\alpha_i.
\end{align}
% Its standard deviation is a known constant. Therefore, we only need $p_{\phi}(x_{T} | \vz)$ to predict the mean of $q(x_T|\vx^0)$, which is $\sqrt{\bar{\alpha}_t} \vx^0$.
% We further let the mean predicted by $p_{\phi}(x_{T} | \vz)$ to be
% \begin{align}
%     \mu = \sqrt{\bar{\alpha}_t} f_{\phi} (\vz).
% \end{align}
Correspondingly, we choose to set $p_{\phi}(\vx^{T'} | \vz)$ to
\begin{align}
\label{eqn:sample_xT}
    p_{\phi}(\vx^{T'} | \vz) \sim N(\vx^{T'}; \sqrt{\bar{\alpha}_t} \vf_{\phi} (\vz), (1-\bar{\alpha}_t)\mI).
\end{align}
Then 
\begin{align}
    D_{\text{KL}} (q(\vx^{T'}|\vx^0) || p_{\phi}(\vx^{T'} | \vz)) = C_1 ||\vx^0-\vf_{\phi} (\vz)||^2 + C_2,
\end{align}
where $C_1, C_2$ are some scalar constants.
Now the second term in $L_{\text{VAE}}$ means that the decoder $\vf_{\phi} (\vz)$ need to reconstruct the original clean image $\vx^0$, based on the latent code $\vz$ sampled from the encoder $q_{\psi}(\vz | \vx^0)$. 
This is exactly the same procedure as training a VAE on clean images $\vx^0$ in the dataset.

\paragraph{Sampling Procedure.} After training the ES-DDPM and the VAE separately, we can combine them to form a new generative model. 
We follow Equation~\ref{eqn:generation_process} to generate samples and it is illustrated in Figure~\ref{fig:es_ddpm_vs_ddpm}: 
First sample $\vz$ from the standard Gaussian distribution, then use the decoder of the VAE to generate an image $\vf_{\phi} (\vz)$. 
Next, sample $\vx^{T'}$ from the distribution described in Equation~\ref{eqn:sample_xT}. 
Finally, use the ES-DDPM to sample $\vx^{t-1}$ from $p_{\bm{\theta}}(\vx^{t-1}|\vx^t)$ for $t=T',T'-1,\cdots,1$, and output the generated image $\vx^0$.
% The above theory derivation and sampling procedure are for unconditional generation. 
% Our method could also be applied to accelerate a conditional DDPM. See details in Appendix~\ref{sec:conditional_generation}.

Note that in the above sampling process, we only need to use the decoder $\vf_{\phi} (\vz)$, while the encoder $q_{\psi}(\vz | \vx^0)$ is not used.
This means that we can not only combine the ES-DDPM with a VAE, but also with any generative model that can map the latent code $\vz$ to a clean image, such as a GAN.
Since after a GAN is trained, its generator is pretty much like the decoder of a VAE, except that the generator's encoder is not known, but we simply do not need the encoder to generate samples from the combined model.
Therefore, we can also think of $\vf_{\phi} (\vz)$ as the generator of a trained GAN.
In fact, we will demonstrate that combining an ES-DDPM with a GAN is more effective than combining it with a VAE in later experiments.
For conditional generation tasks, we can also combine a conditional ES-DDPM with a conditional VAE or GAN to achieve acceleration. See details in Appendix~\ref{sec:conditional_generation}.

% \vspace{-0.5em}
\section{Related Works}
% \vspace{-0.5em}
Several works~\citep{ho2020denoising, kong2021fast, nichol2021improved} propose to accelerate the sampling procedure of DDPMs by making jumping steps in the reverse steps.
\citet{watson2021learning} propose a dynamic programming algorithm that finds the optimal denoising time-step schedule for DDPMs.
\citet{watson2021learning} optimize DDPM fast samplers by differentiating through sample quality scores.
\citet{salimans2022progressive} propose to progressively distill a trained DDPM for fast sampling.
\citet{xiao2021tackling} propose denoising diffusion GANs that can balance sample quality, mode coverage, and fast sampling.
We think our method is orthogonal to these acceleration methods.
In principle, these methods can be coupled with our method to accelerate the $T'$ denoising steps in our ES-DDPM.

Some works find that combining DDPMs with VAEs can help accelerate the sampling process of DDPMs. 
\citet{preechakul2021diffusion} condition the reverse process of the DDPM on an encoded vector of an image.
\citet{pandey2022diffusevae} condition the reverse process of the DDPM on an reconstructed image by a VAE.
\citet{kingma2021variational} design variational diffusion models that can generate images in fewer steps than a DDPM.
However, all of them still need a complete diffusion process to destroy the input image, and in their sampling process, they still need to run the full denoising process that starts with a Gaussian noise or a very noisy image.
In comparison, we adopt early stop in the diffusion process, and we start from a partially diffused image generated by a VAE or GAN in our denoising process. 
Another major difference is that these methods need to alter the training process and train a DDPM that is specific to the VAE model they want to combine with, while our method can combine a pre-trained DDPM with any other generative models.

% \textcolor{red}
We find that a concurrent work~\citep{zheng2022truncated} adopts a similar acceleration method to our ES-DDPM,
where ES-DDPM represents the non-Gaussian distribution of $\vx^{T'}$ by diffusing clean samples obtained from a pre-trained generative model, while the concurrent work proposes to directly learn the distribution of $\vx^{T'}$ with a GAN or a conditional transport.
Consequently, ES-DDPM bears the following advantages:
%But different from our method, they propose to utilize a GAN or conditional transport~\citep{zheng2021exploiting} to directly learns the distribution of $\vx^{T'}$, 
%while we propose to obtain $\vx^{T'}$ by diffusing clean images generated by other generative models such as VAEs or GANs. 
%We think that our method bears several advantages: 
First, by diffusing the real data distribution approximated by a pre-trained generative model, ES-DDPM avoids learning  the distribution of $\vx^{T'}$ directly, 
which brings extra computational complexity,
and learning the distribution of noisy images $\vx^{T'}$ may be harder than learning the real data distribution.
%need the GAN or VAE to spend extra capacity on learning the noise distribution in $\vx^{T'}$. 
%They only need to model the distribution of the clean images, and the noises can be added afterwards by hand.
% We compare the sample quality and acceleration effect of our method with theirs in our experiments.
Second, ES-DDPM can use the same pre-trained generative model for any value of $T'$, while the concurrent work requires to learn a new model for each possible value of $T'$.
%one generative model trained on clean images for any $T'$, while theirs need to train a new generative model for every $T'$.
Third, without re-training, ES-DDPM can easily utilize the widely-available pre-trained generative models in the community, and seamlessly integrate advances in the related fields. In this way, attractive properties, such as controllable generation, possessed by existing generative models can be effectively introduced to DDPMs.
%Third, our method enables us to utilize the widely-available pre-trained generative models (trained on clean images) such as GANs and VAEs in the community. 
%Many properties such as controllable generation are %extensively studied for these pre-trained generative models and can be utilized in our combined model.
%We compare their method with ours in the experiments.

% \vspace{-0.5em}
\section{Experiments}
% \vspace{-0.5em}
\label{sec:experiments}
\subsection{Image Generation}
\label{sec:image_generation}
In this section, we evaluate the image generation performance of our ES-DDPM and compare it with other generative models.

\vspace{-0.5em}
\paragraph{Datasets.} We conduct experiments on CIFAR-10~\citep{krizhevsky2009learning}, CelebA~\citep{liu2015faceattributes}, ImageNet~\citep{russakovsky2015imagenet}, LSUN-Bedroom and LSUN-Cat~\citep{yu2015lsun}.
For the CelebA dataset, we conduct experiments at two resolutions: $64\times 64$ and $128\times 128$. For the ImageNet dataset, we use a resolution of $64\times 64$, and we use $256\times 256$ for the LSUN datasets.

\vspace{-0.5em}
\paragraph{Evaluation Metrics.} 
We use the commonly used Fréchet inception distance~\citep{heusel2017gans} (FID) and Inception Score~\citep{salimans2016improved} (IS) to evaluate the quality of generated images.
We also use Improved Precision and Recall metrics~\citep{kynkaanniemi2019improved} to explicitly measure sample fidelity and diversity.
sFID~\citep{nash2021generating} is also used.

\vspace{-0.5em}
\paragraph{CIFAR-10 Experiment.}
We combine ES-DDPM with other generative models, \ie, vannila VAE~\citep{kingma2013auto}, DCGAN~\citep{radford2015unsupervised}, StyleGAN2~\citep{karras2020analyzing}, and test their performance in terms of unconditional image generation on CIFAR-10~\citep{krizhevsky2009learning} dataset.
We use the pre-trained 1000-step DDPM on CIFAR-10 from the work~\citep{ho2020denoising}.
We insert the images generated by the other generative model (VAE or GAN) to different steps in the reverse process of the DDPM, namely, use different $T'$ in Equation~\ref{eqn:generation_process}.
We report FID of the combined generative models in Table~\ref{tbl:cifar10}.

We can see that DDPM can improve the quality of images generated by the other generative models.
And roughly speaking, the larger the number of denoising steps $T'$ is, the better the quality is. 
This is because $p_{\phi}(\vx^{T'} | \vz)$ in Equation~\ref{eqn:sample_xT} gradually approaches to a standard Gaussian noise as $T'$ increases, and hence the combined model gradually matches the original DDPM as $T'$ increases.
Another observation is that we need to combine ES-DDPM with a strong generative model like StyleGAN2 if we want to obtain high quality images in few denoising steps.
One rather surprising observation is that the combined model can even outperform the original DDPM.
In other words, we can combine a DDPM with a less powerful generative model, and with less denoising steps, to obtain a even better combined model, which means we can accelerate the DDPM for free!
As we will show in later experiments, the same results can be achieved on other datasets as well.
% We also compare our method with DiffuseVAE~\citep{pandey2022diffusevae} in Table~\ref{tbl:cifar10}.
% We can see that our method can achieve better generation quality than DiffuseVAE using the same number of denoising steps when combined with StyleGAN2.

\vspace{-0.5em}
\paragraph{CelebA Experiment.} We combine ES-DDPM with StyleGAN2 and test its performance on the CelebA~\citep{liu2015faceattributes} dataset at two resolutions: CelebA-64 and CelebA-128.
At each resolution, we train two DDPMs with $T'=100$ and $T'=200$ using the loss function in Equation~\ref{eqn:l_ddpm}, and combine them with a self-trained  StyleGAN2, respectively.
We compare our combined models with self-trained $1000$-step DDPMs and other generative models in Table~\ref{tbl:celeba_64}.
The DDPMs are trained with a linear $\beta$ schedule following the same procedures in the works~\citep{ho2020denoising, song2020denoising}.
We can see that when combined with StyleGAN2, our ES-DDPM can achieve lower FIDs than the original DDPM and other score-based methods in even less denoising steps.
\begin{table}[t]
% \parbox{0.49\columnwidth}{ 
\centering
\vspace{-2em}
\caption{Unconditional image generation performance on CIFAR-10. 
We report FID between randomly generated $50000$ images by the combined generative model and the training set.
$T'=0$ refers to the original VAE or GAN model.
$T'=1000$ is the original 1000-step DDPM.}
\label{tbl:cifar10}
\resizebox{1\columnwidth}{!}{
% \small{
\begin{tabular}{c|c|c|c|c|c|c|c|c|c|c|c}
    \Xhline{1pt}
	Denoising Steps $T'$ & 0 & 100 & 200 & 300 & 400 & 500 & 600 & 700 & 800 & 900 & 1000 \\
	\hline \hline
	VAE+ES-DDPM  & 158.61 & 49.62 & 20.88 & 11.03 & 5.96 & 3.69 & 3.17 & 3.15 & \textbf{3.12} & 3.18 & \multirow{3}{*}{3.20}\\ 
% 	\cline{1-10}
	DCGAN+ES-DDPM & 33.31 & 15.54 & 9.97 & 7.21 & 5.49 & 3.98 & 3.37 & \textbf{3.13} & 3.14 & 3.24\\
% 	\cline{1-10}
	StyleGAN2+ES-DDPM & 7.18 & 5.52 & 5.02 & 4.60 & 4.03 & 3.51 & 3.26 & \textbf{3.11} & 3.16 & 3.17\\
% 	\cline{1-12}
% 	DiffuseVAE~\citep{pandey2022diffusevae} &&11.71&&&&9.72&&&&&8.72 \\
	\Xhline{1pt}
\end{tabular}
% }
}
% }
\end{table}
% \begin{table}[thb]
% % \parbox{0.49\columnwidth}{ 
% \centering
% \caption{Celeba-64 unconditional image generation}
% \label{tbl:cifar10}
% % \resizebox{1\columnwidth}{!}{
% % \small{
% \begin{tabular}{c|c|c|c|c|c|c|c|c|c|c|c}
%     \Xhline{1pt}
% 	Celeba-64 & 0 & 100 & 200 & DDPM \\
% 	\hline \hline
% 	FID       & 4.55 & 3.01 & \textbf{2.55} & 3.26\\
% % 	\cline{1-10}
% 	IS        & 2.64 & \textbf{2.65} & 2.62 & 2.39\\ 
% % 	\cline{1-10}
% 	\Xhline{1pt}
% \end{tabular}
% % }
% % }
% % }
% \end{table}

\begin{table}[t]
% \parbox{0.49\columnwidth}{ 
\centering
\caption{Performance comparison of CelebA unconditional image generation. ``*'' means the model is trained by ourselves. 
% and images for evaluation are generated by ourselves. It has the same meaning in latter tables as well. 
For our method and self trained models, we report FID between randomly generated $50000$ images and the whole dataset.}
\label{tbl:celeba_64}
\resizebox{0.7\columnwidth}{!}{
% \small{
\begin{tabular}{ccc}
    \Xhline{1pt}
	CelebA-64 & Method & FID \\
	\hline \hline
	\multirow{3}{*}{Hybrid Models} & StyleGAN2+$100$-step ES-DDPM (\textbf{Ours}) & 3.01\\
	& StyleGAN2+$200$-step ES-DDPM (\textbf{Ours}) & \textbf{2.55}\\
% 	& $100$-step TDPM-GAN~\citep{zheng2022truncated} & 3.64 \\
	& $1000$-step DiffuseVAE~\citep{pandey2022diffusevae} & 4.76\\
	\hline
	\multirow{4}{*}{Score-based Methods} & 1000-step DDPM*~\citep{ho2020denoising} & 3.26\\
	& $250$-step PNDM~\citep{liu2022pseudo} & 2.71\\
	& NCSN~\citep{song2019generative} & 25.30\\
	& NCSNv2~\citep{song2020improved} & 10.23\\
	\hline
	\multirow{3}{*}{GAN-based Methods} & COCO-GAN~\citep{lin2019coco} & 4.00\\
	& StyleGAN2*~\citep{karras2020analyzing} & 4.55\\
	& QA-GAN~\citep{parimala2019quality} & 6.42\\
	\hline
	VAE-based Methods & NCP-VAE~\citep{aneja2020ncp} & 5.25\\
	\Xhline{0.7pt}
	\\
	\Xhline{0.7pt}
	CelebA-128 & Method & FID \\
	\hline \hline
	\multirow{2}{*}{Hybrid Models} & StyleGAN2+$100$-step ES-DDPM (\textbf{Ours}) & \textbf{1.76}\\
	& StyleGAN2+$200$-step ES-DDPM (\textbf{Ours}) & 1.79\\
	\hline
	\multirow{1}{*}{Score-based Methods} & 1000-step DDPM*~\citep{ho2020denoising} & 5.65\\
	\hline
	\multirow{3}{*}{GAN-based Methods} & COCO-GAN~\citep{lin2019coco} & 5.74\\
	& StyleGAN2*~\citep{karras2020analyzing} & 2.13\\
	& PresGAN~\citep{dieng2019prescribed} & 29.12\\
	\Xhline{1pt}
\end{tabular}
% }
}
% }
\vspace{-1em}
\end{table}

% \begin{table}[thb]
% % \parbox{0.49\columnwidth}{ 
% \centering
% \caption{Celeba-128 unconditional image generation}
% \label{tbl:cifar10}
% % \resizebox{1\columnwidth}{!}{
% % \small{
% \begin{tabular}{c|c|c|c|c|c|c|c|c|c|c|c}
%     \Xhline{1pt}
% 	Celeba-128 & 0 & 100 & 200 & DDPM \\
% 	\hline \hline
% 	FID       & 2.13 & \textbf{1.76} & 1.79 & 5.65\\
% % 	\cline{1-10}
% 	IS        & \textbf{3.03} & 2.98 & 2.96 & 2.49\\ 
% % 	\cline{1-10}
% 	\Xhline{1pt}
% \end{tabular}
% % }
% % }
% % }
% \end{table}

Compared with CIFAR-10 experiment results in Table~\ref{tbl:cifar10}, we find that ES-DDPM combined with StyleGAN2 can outperform the full-length DDPM in much less denoising steps on the CelebA dataset.
In Table~\ref{tbl:celeba_64}, $T'=100$ or $T'=200$ can already outperform the $1000$-step DDPM, while in Table~\ref{tbl:cifar10}, we need $700$ to $800$ denoising steps to achieve a lower FID than the full-length DDPM.
We think it is because CIFAR-10 is a multi-category dataset while CelebA is a single-category dataset. In other words, CIFAR-10 has more data diversity than CelebA.
GANs are known to be able to generate high quality samples, but have poor mode coverage~\citep{zhao2018bias, xiao2021tackling}.
On the other hand, DDPMs demonstrate both high sample quality and good mode coverage~\citep{dhariwal2021diffusion}.
Therefore, for an ES-DDPM to improve the generation diversity of a GAN on a dataset of complex distribution, we need large $T'$s in Equation~\ref{eqn:partial_reverse_process}.
Because for small $T'$s, the ES-DDPM generated image $\vx^0$ will be very similar to the GAN generated image $\vf_{\phi} (\vz)$, as there is little diversity and stochasticity in Equation~\ref{eqn:partial_reverse_process} for small $T'$s. 
On the other hand, for large $T'$s, the generated image $\vx^0$ can be vastly different from $\vf_{\phi} (\vz)$, because $\vx^{T'}$ contains very little information of $\vf_{\phi} (\vz)$, and thus the GAN's generation diversity can be improved.
We will explicitly verify this point of view using the Precision and Recall metrics in the following ImageNet and LSUN Experiments.

\paragraph{ImageNet and LSUN Experiment.} For ImageNet-64, we combine ES-DDPM with BigGAN-deep~\citep{brock2018large} and test its class-conditional image generation performance.
For LSUN-Bedroom and LSUN-Cat, we combine ES-DDPM with StyleGAN~\citep{karras2019style} and StyleGAN2, respectively.
For these experiments, we use pre-trained DDPM models and GAN generated images provided in the work~\citep{dhariwal2021diffusion}.
The DDPMs are of length $1000$ and the variance schedule is learned by a neural network.
We follow the same evaluation methods in the work~\citep{dhariwal2021diffusion}:
FIDs are computed between randomly generated $50000$ images and the whole training set.
IS's are computed over the randomly generated $50000$ images. Precision and Recall are computed between the randomly generated $50000$ images and random $10000$ images from the training set.

Experiment results are shown in Table~\ref{tbl:imagenet_64}, Table~\ref{tbl:lsun_bedroom} and Table~\ref{tbl:lsun_cat}.
Same as CIFAR-10 and CelebA, we can see that ES-DDPM can improve the generation quality of GANs, and the combined model can outperform the original DDPM in terms of FID, sFID and IS.
We also observe that BigGAN-deep has low Recall on the complex multi-category ImageNet dataset, its Recall is gradually improved as $T'$ increases. Therefore, the combined model need large $T'$s around $900$ to achieve a lower FID than the full-length DDPM.
On the other hand, for the single-category datasets, LSUN-Bedroom and LSUN-Cat, StyleGAN and StyleGAN2 already have a not too bad Recall compared with the full-length DDPMs. Therefore, the ES-DDPMs can improve the GANs' Recall in less denoising steps, and our combined model can outperform the original full-length DDPM in terms of FID in just $100$ steps.
In conclusion, for multi-category datasets of complex data distribution, our combined models provide a trade-off between sample diversity and denoising steps. 
For single-category datasets where GANs already have a good Recall, our combined models provide better sample quality than both the GANs and the original full-length DDPMs, with much fewer denoising steps than the original full-length DDPMs.
% \begin{table}[thb]
% % \parbox{0.49\columnwidth}{ 
% \centering
% \caption{Imagenet-64 class conditional image generation FID. DDPM is sampled with classifier guidance}
% \label{tbl:cifar10}
% \resizebox{1\columnwidth}{!}{
% % \small{
% \begin{tabular}{c|c|c|c|c|c|c|c|c|c|c|c}
%     \Xhline{1pt}
% 	Reverse steps & 0 & 100 & 200 & 300 & 400 & 500 & 600 & 700 & 800 & 900 & DDPM \\
% 	\hline \hline
% 	FID       & 4.06  & 3.71  & 3.47  & 3.21  & 2.97  & 2.75  & 2.54  & \textbf{2.33}  & 2.37  & 2.82  & 4.16\\
% % 	\cline{1-10}
% 	sFID      & 3.96  & 3.91  & 3.93  & 3.93  & 3.87  & 3.83  & \textbf{3.82}  & \textbf{3.82}  & 3.86  & 4.22  & 5.73\\
% % 	\cline{1-10}
% 	IS        & 45.00 & 49.3  & 51.39 & 53.72 & 56.03 & 58.98 & 63.53 & 68.70 & 74.34 & 83.28 & \textbf{92.40}\\ 
% % 	\cline{1-10}
% 	Precision & 0.795 & 0.799 & 0.802 & 0.798 & 0.799 & 0.797 & 0.798 & 0.803 & 0.816 & 0.822 & \textbf{0.837}\\
% % 	\cline{1-10}
% 	Recall    & 0.483 & 0.495 & 0.511 & 0.527 & 0.536 & 0.543 & 0.549 & 0.558 & \textbf{0.563} & 0.552 & 0.532\\
% 	\Xhline{1pt}
% \end{tabular}
% % }
% }
% % }
% \end{table}

\begin{table}[t]
% \parbox{0.49\columnwidth}{ 
\centering
\vspace{-4em}
\caption{\small{ImageNet-64 class-conditional image generation performance of ES-DDPM+BigGAN-deep. 
For the DDPM and ES-DDPM experiments, we follow the work~\citep{dhariwal2021diffusion} to use Timestep-Respacing with jumping interval $4$. The actual denoising steps is $T'/4$. We do not use classifier guidance.}}
\label{tbl:imagenet_64}
\resizebox{1\columnwidth}{!}{
% \small{
\begin{tabular}{c|c|c|c|c|c|c|c|c|c|c|c}
    \Xhline{1pt}
	ES-DDPM Length $T'$ & 0(BigGAN-deep) & 100 & 200 & 300 & 400 & 500 & 600 & 700 & 800 & 900 & 1000(DDPM) \\
	\hline \hline
	FID       & 4.06  & 3.75  & 3.47  & 3.30  & 3.16  & 2.92  & 2.71  & 2.50  & 2.31  & \textbf{2.07}  & 2.13\\
% 	\cline{1-10}
	sFID      & 3.96  & 3.91  & 3.96  & 3.93  & 3.90  & \textbf{3.83}  & 3.84  & 3.84  & 3.84  & 3.90  & 4.28\\
% 	\cline{1-10}
	IS        & 45.00 & 48.63 & 50.17 & 51.45 & 52.32 & 53.85 & 54.41 & 55.11 & \textbf{55.42} & 55.29 & 52.52\\ 
% 	\cline{1-10}
	Precision & 0.795 & 0.798 & \textbf{0.800} & 0.794 & 0.792 & 0.787 & 0.784 & 0.778 & 0.771 & 0.756 & 0.739\\
% 	\cline{1-10}
	Recall    & 0.483 & 0.504 & 0.514 & 0.529 & 0.533 & 0.547 & 0.565 & 0.577 & 0.593 & 0.608 & \textbf{0.631}\\
	\Xhline{1pt}
\end{tabular}
% }
}
% }
\end{table}
\begin{table}[t]
% \vspace{-1em}
\parbox{0.49\columnwidth}{ 
\centering
\caption{\small{ES-DDPM+StyleGAN on LSUN-Bedroom.}}
\label{tbl:lsun_bedroom}
\resizebox{0.49\columnwidth}{!}{
% \small{
\begin{tabular}{c|c|c|c|c|c}
    \Xhline{1pt}
	Denoising & 0 & \multirow{2}{*}{100} & \multirow{2}{*}{200} & \multirow{2}{*}{300} & 1000 \\
	Steps $T'$ & (StyleGAN) & & & & (DDPM)\\
	\hline \hline
	FID       & 2.35  & 1.85  & 1.70  & \textbf{1.68}  & 1.86\\
% 	\cline{1-10}
	sFID      & 6.61  & 5.93  & 5.70  & 5.75  & \textbf{5.64}\\
% 	\cline{1-10}
	IS        & \textbf{2.55}  & 2.53  & 2.54  & 2.54  & 2.39\\ 
% 	\cline{1-10}
	Precision & 0.590 & 0.636 & 0.642 & 0.643 & \textbf{0.653}\\
% 	\cline{1-10}
	Recall    & 0.483 & 0.464 & 0.472 & 0.495 & \textbf{0.500}\\
	\Xhline{1pt}
\end{tabular}
% }
}
}
\parbox{0.49\columnwidth}{ 
\centering
\caption{\small{ES-DDPM+StyleGAN2 on LSUN-Cat.}}
\label{tbl:lsun_cat}
\resizebox{0.49\columnwidth}{!}{
% \small{
\begin{tabular}{c|c|c|c|c|c}
    \Xhline{1pt}
	Denoising & 0 & \multirow{2}{*}{100} & \multirow{2}{*}{200} & \multirow{2}{*}{300} & 1000 \\
	Steps $T'$ & (StyleGAN2) & & & & (DDPM)\\
	\hline \hline
	FID       & 7.26  & 5.47  & 5.15  & \textbf{4.89}  & 5.68\\
% 	\cline{1-10}
	sFID      & \textbf{6.33}  & 6.47  & 6.57  & 6.70  & 6.72\\
% 	\cline{1-10}
	IS        & 4.84  & 5.12  & 5.23  & \textbf{5.33}  & 5.23\\ 
% 	\cline{1-10}
	Precision & 0.576 & 0.626 & \textbf{0.633} & 0.631 & 0.628\\
% 	\cline{1-10}
	Recall    & 0.432 & 0.463 & 0.475 & 0.489 & \textbf{0.520}\\
	\Xhline{1pt}
\end{tabular}
% }
}
}
\vspace{-1em}
\end{table}

\vspace{-0.25em}
\subsection{Coupled With Other DDPM Acceleration Methods}
\vspace{-0.25em}
Our method can be easily coupled with other DDPM acceleration methods to achieve further acceleration.
Recall that our method need to use the ES-DDPM to denoise $x^{T'}$ for $T'$ steps as shown in Equation~\ref{eqn:partial_reverse_process}.
This denoising process is actually the same as the last $T'$ steps of the reverse process in a normal full-length DDPM.
Any method that aims at accelerating the full $T$-step reverse process of a DDPM, can in principle be applied to accelerate the last $T'$ denoising steps. 
We use DDIM~\citep{song2020denoising} and Timestep-Respacing (TR)~\citep{nichol2021improved}  as a demonstration in our experiments.
Both DDIM and Timestep-Respacing enable one to sampling from a pre-trained DDPM only on a subsequence of the original denoising steps.
We choose a uniform subsequence in our $T'$ denoising steps. 
We conduct experiments on the CelebA, ImageNet, LSUN-bedroom and LSUN-Cat datasets. 
CelebA and ImageNet results are shown in Table~\ref{tbl:celeba_acceleration}, Table~\ref{tbl:imagenet_64_acceleration}.
LSUN-bedroom and LSUN-Cat results are shown in
Table~\ref{tbl:lsun_bedroom_acceleration} and Table~\ref{tbl:lsun_cat_acceleration} in Appendix~\ref{sec:lsun_ddpm_acceleration}.
We can see that our combined model can indeed achieve further acceleration when coupled with other acceleration methods.
And compared with other acceleration methods, our combined model bears better sample quality when using the same number of denoising steps in most cases.

% We combine StyleGAN2 with the DDIM accelerated DDPM and compare it with other acceleration methods in Table~\ref{tbl:celeba_acceleration}.

% And we think our method could achieve even better results when coupled with more advanced acceleration methods.
% \begin{table}[thb]
% % \parbox{0.49\columnwidth}{ 
% \centering
% \caption{Celeba-64 unconditional image generation. DDIM acceleration of the original 100 step ddpm.}
% \label{tbl:cifar10}
% % \resizebox{1\columnwidth}{!}{
% % \small{
% \begin{tabular}{c|c|c|c|c|c|c|c|c|c|c|c}
%     \Xhline{1pt}
%     skip step    & 2  & 5  & 10 & 20 \\
%     Actual steps & 50 & 20 & 10 & 5 \\
% 	\hline \hline
% 	FID      & 3.97 & 4.90 & 6.44 & 9.15\\
% % 	\cline{1-10}
% 	IS       & 2.69 & 2.69 & 2.69 & 2.69\\ 
% % 	\cline{1-10}
% 	\Xhline{1pt}
% \end{tabular}
% % }
% % }
% % }
% \end{table}

\begin{table}[thb]
% \parbox{0.49\columnwidth}{ 
\centering
\vspace{-4em}
\caption{FID comparison of different DDPM acceleration methods on CelebA. For our models, denoising steps less than $100$ steps are achieved by accelerating ES-DDPM($T'=100$) using DDIM.}
\label{tbl:celeba_acceleration}
\resizebox{0.9\columnwidth}{!}{
% \small{
\begin{tabular}{c|c|c|c|c|c|c|c}
    \Xhline{1pt}
    & Denoising Steps & 200 & 100 & 50 & 20 & 10 & 5 \\
	\hline \hline
	\multirow{6}{*}{CelebA-64} & ES-DDPM+StyleGAN2+DDIM(\textbf{Ours})  & \textbf{2.55}  & 3.01 & 3.97 & \textbf{4.90} & \textbf{6.44} & \textbf{9.15}\\
% 	\hline
    & DDIM~\citep{song2020denoising}& - & 6.53 & 9.17 & 13.73 & 17.33 & - \\
	& FastDPM~\citep{kong2021fast}& - & 7.85 & 8.31 & 10.69 & 15.31 & - \\
% 	\hline
	& PNDM~\citep{liu2022pseudo}& 2.71 & \textbf{2.81} & 3.34 & 5.51 & 7.71 & 11.3 \\
	& Diffusion Autoencoder~\citep{preechakul2021diffusion} & - & 5.30 & 7.05 & 10.18 & 12.92 & - \\
	& TDPM-GAN~\citep{zheng2022truncated} & - &3.64&\textbf{3.28}& - & - & - \\
	\hline
	\multirow{1}{*}{CelebA-128} & ES-DDPM+StyleGAN2+DDIM(\textbf{Ours})  & 1.79 & 1.76 & 1.86 & 2.27 & 3.26 & 6.15\\
% 	 & DDIM*~\citep{song2020denoising}  & 26.24 & 26.32 & 29.30 & 37.94 & 41.84 & 49.83\\
	\Xhline{1pt}
\end{tabular}
% }
}
% }
\end{table}

% \begin{table}[thb]
% % \parbox{0.49\columnwidth}{ 
% \centering
% \caption{Celeba-128 unconditional image generation. DDIM acceleration of the original 100 step ddpm.}
% \label{tbl:cifar10}
% % \resizebox{1\columnwidth}{!}{
% % \small{
% \begin{tabular}{c|c|c|c|c|c|c|c|c|c|c|c}
%     \Xhline{1pt}
%     skip step    & 2  & 5  & 10 & 20 \\
%     Actual steps & 50 & 20 & 10 & 5 \\
% 	\hline \hline
% 	FID      & 50 & 2.27 & 3.26 & 6.15\\
% % 	\cline{1-10}
% 	IS       & 50 & 3.02 & 3.04 & 3.07\\ 
% % 	\cline{1-10}
% 	\Xhline{1pt}
% \end{tabular}
% % }
% % }
% % }
% \end{table}

% Next, we couple our acceleration method with DDIM and Timestep-Respacing~\citep{nichol2021improved}. Results are shown in Table~\ref{tbl:imagenet_64_acceleration}.
% We can see that in most cases, our method achieves lower FIDs and higher IS's using the same number of denoising steps.
\begin{table}[thb]
% \parbox{0.49\columnwidth}{ 
\centering
\caption{FID and IS comparison of different DDPM acceleration methods on the Imagenet-64 dataset. ``*'' means the images for evaluation are generated by ourselves for this method. }
% For our models, acceleration is achieved by coulping ES-DDPM($T'=100$) with Timespace-Respacing.}
\label{tbl:imagenet_64_acceleration}
\resizebox{1\columnwidth}{!}{
% \small{
% \begin{tabular}{c|c|c|c|c|c|c|c|c|c|c|c| c|c|c|c|c}
\begin{tabular}{c|ccccc | ccccc}
    \Xhline{1pt}
    & \multicolumn{5}{c|}{FID} & \multicolumn{5}{c}{IS}\\
    \hline
	Denoising Steps       & 25 & 20 & 15 & 10 & 5 & 25 & 20 & 15 & 10 & 5\\
	\hline \hline
    \small{ES-DDPM($T'=100$)+BigGAN-deep+TR(\textbf{Ours})} & \textbf{3.75} & \textbf{3.77} & \textbf{3.80} & \textbf{3.93} & \textbf{4.25} & 48.63 & \textbf{48.92} & \textbf{49.17} & \textbf{48.81} & \textbf{48.04}\\
% 	\cline{1-10}
	DDIM*~\citep{song2020denoising} & 5.90 & 6.68 & 8.59 & 13.48 & 67.70 & 38.24 & 36.72 & 34.85 & 30.79 & 13.17\\
% 	\cline{1-10}
	TR*~\citep{nichol2021improved}  & 7.99 & 11.08 & 16.49 & 27.74 & 78.59 & \textbf{49.62} & 45.83 & 39.32 & 29.32 & 12.65\\ 
% 	\cline{1-10}
    DDSS~\citep{watson2021learning} & 18.40 & 20.69 & 24.69 & 37.32 & 55.14& 18.12 & 17.92 & 17.23 & 14.76 & 12.90\\
	\Xhline{1pt}
\end{tabular}
% }
}
% }
\vspace{-1em}
\end{table}

\vspace{-0.25em}
\subsection{Accelerate Training}
\begin{figure}[thb]
    % \centering
    \vspace{-0.8em}
    \subfigure[CelebA-64]{
    % \label{fig:mlp}
    \includegraphics[width=0.483\textwidth]{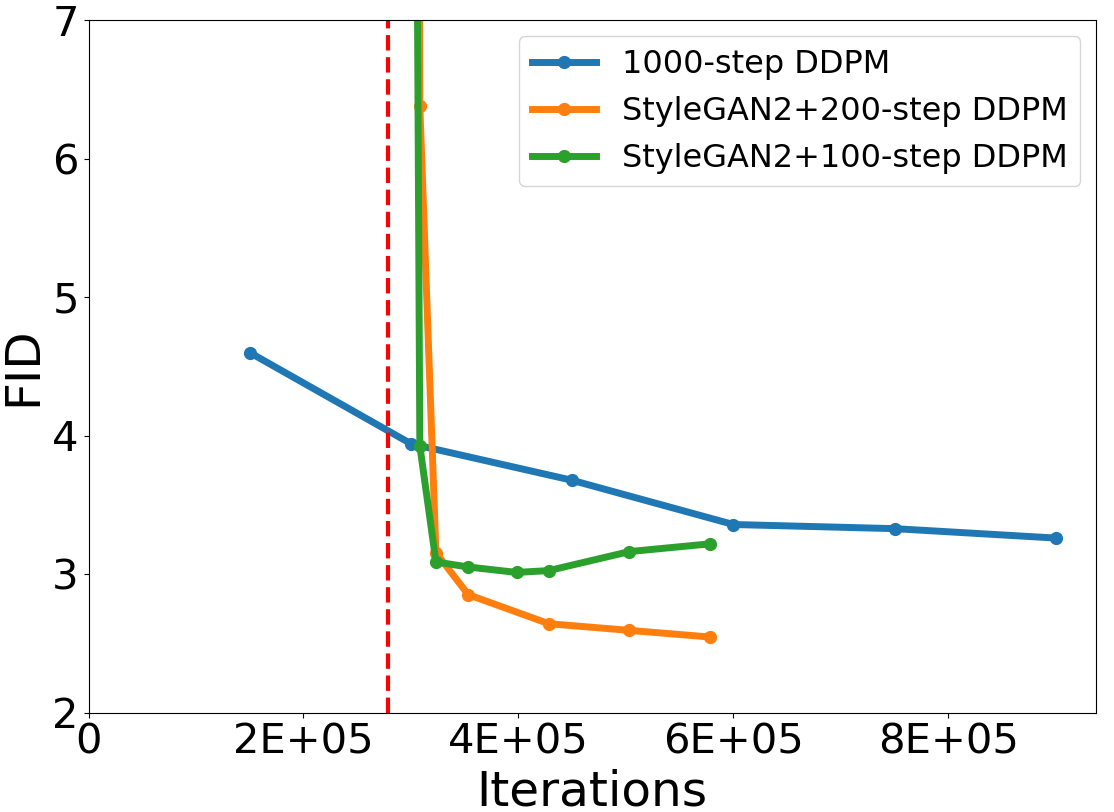}}
    \hspace{0.01\textwidth}
    \subfigure[CelebA-128]{
    % \label{fig:local_feature}
    \includegraphics[width=0.485\textwidth]{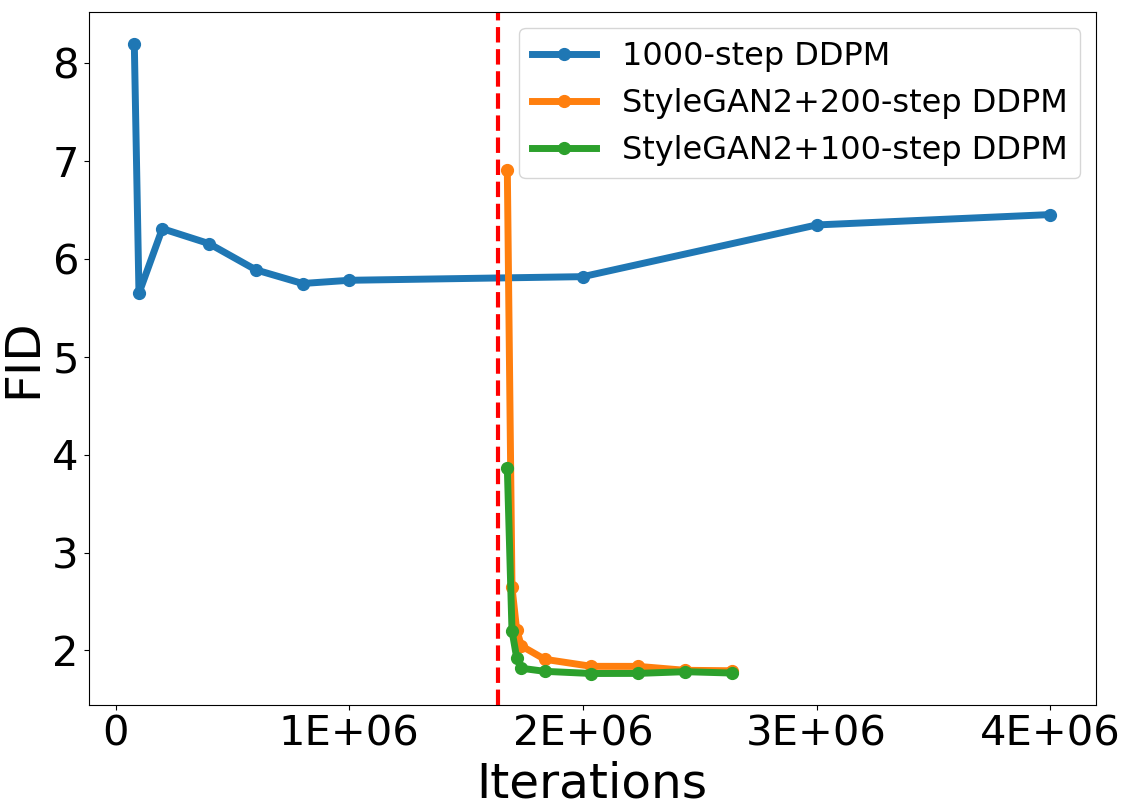}}
    \vspace{-1em}
    \caption{Training time comparison between our combined models and the original full-length DDPMs.
    DDPMs are trained with batchsize $128$ on CelebA-64 and with batchsize $48$ on CelebA-128.
    The red dashed line denotes the training time of StyleGAN2.}
    \label{fig:celeba_accelerate_training}
    \vspace{-1em}
\end{figure}

\vspace{-0.25em}
We have shown that our combined model can accelerate the sampling process of DDPMs in the previous sections.
We also expect our method to be able to accelerate the training process of DDPMs, since we only need to train a denoising chain of length $T'$, which is shorter than the original denoising chain of length $T$.
This is because our DDPM loss in Equation~\ref{eqn:l_ddpm} contains only $T'$ terms, compared with $T$ terms in the original full-length DDPM loss.
We compare the training process of our combined models with the full-length $1000$-step DDPM on the CelebA dataset.
We plot FIDs during the training process of our combined model and the original DDPM in Figure~\ref{fig:celeba_accelerate_training}.
Note that we have already taken into account the training time of the StyleGAN2 by converting the training time of  the StyleGAN2 to the equivalent training steps of the DDPM, and it is annotated by a red dashed line in Figure~\ref{fig:celeba_accelerate_training}.
For CelebA-64, we can see that our combined model still converges faster than the full-length DDPM even if we take into account the training time of the StyleGAN2.
Specifically, it takes the full-length DDPM $9\times10^5$ iterations, about $80$ hours to achieve an FID of $3.26$.
In comparison, our ES-DDPM($T'=100$) only uses $4.5\times10^4$ iterations, about $4$ hours to achieve an FID of $3.09$.
Even if we take into account the training time of StyleGAN2, which is about $25$ hours, the total training time of our combined model ($29$ hours) is still much shorter than the training time of the full-length DDPM ($80$ hours).
As for CelebA-128, we find the full-length DDPM could not achieve an FID comparable to our combined model even we train it for a very long time.

\subsection{Controllable Image Generation}
\begin{figure}[thb]
    % \centering
    \vspace{-4em}
    % \label{fig:mlp}
    \includegraphics[width=1\textwidth]{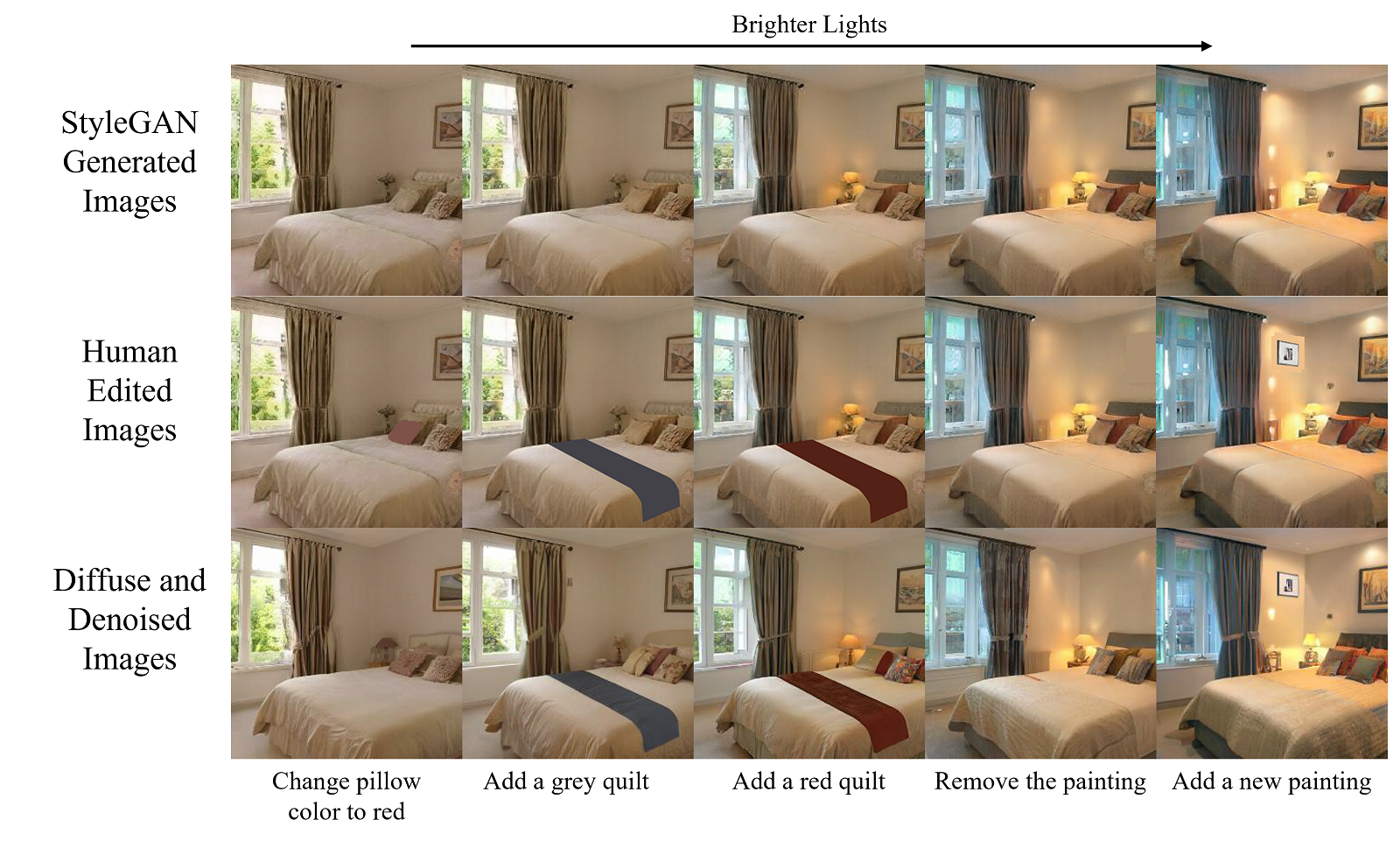}
    \caption{On LSUN-Bedroom-256, we use StyleGAN to control the lighting condition of a bedroom, and then use the ES-DDPM($T'=200$) to control local details.}
    \label{fig:bedroom_editing}
    \vspace{-1em}
\end{figure}

In previous sections, we have shown that combining GANs and DDPMs can improve their image generation quality, and accelerate the sampling and training process of DDPMs.
In this section, we further show that the combined models can benefit from both GANs and DDPMs in terms of controllable generation.
DDPMs have achieved impressive results on image local editing~\citep{nichol2021glide, meng2021sdedit}. 
We think this can be largely attributed to DDPMs' latent space $\vx^t$, which has the same dimension as the input image and thus allows us to perform pixel-level manipulation on this latent space.
However, this latent space lack semantic meanings to control high-level semantics of the generated image.
On the other hand, GANs and VAEs can learn a meaningful representation of an image that can be used to control high-level semantics, but it's challenging for them to achieve pixel-level control of the generated image.
We think that DDPMs and GANs (or VAEs) are complementary to each other in terms of controllable generation.
We can achieve both high-level semantic control and pixel-level control of the generated image in our combined models.

It is achieved by following Equation~\ref{eqn:generation_process} to generate images: 
First use a GAN or VAE to generate an image $\vf_{\phi} (\vz)$. In this step, the GAN or VAE can control the high-level semantics of the image. 
Next, we can perform some coarse local editing on the image. 
Then, sample $\vx^{T'}$ from the distribution $N(\vx^{T'}; \sqrt{\bar{\alpha}_t} \vf_{\phi} (\vz), (1-\bar{\alpha}_t)\mI)$. 
Finally, use the ES-DDPM to denoise $\vx^{T'}$ by sampling $\vx^{t-1}$ from $p_{\bm{\theta}}(\vx^{t-1}|\vx^t)$ for $t=T',T'-1,\cdots,1$, and output the generated image $\vx^0$.
The denoising process will preserve high-level semantics in $\vf_{\phi} (\vz)$ as long as $T'$ is not too large, and it will refine the local editing we performed and mitigate artifacts in the image.

We conduct experiments to verify the effect of our method on controllable generation.
Figure~\ref{fig:bedroom_editing} shows that we can use StyleGAN to control the lighting condition of a bedroom, and then perform local editing on the generated images (similar to the stroke-based editing in the work~\citep{meng2021sdedit}). 
Finally, we use the ES-DDPM to refine the edited images, mitigate artifacts and make the images photo-realistic.
In this way, our combined model can control both the high-level lighting condition and local details of the bedroom.
Additional examples are provided in Appendix Section~\ref{sec:additional_controllable_generation_exps}.

\section{Conclusion}
\label{sec:conclusion}
In this work, we propose to adopt early stop in the diffusion process of a DDPM to achieve acceleration.
We can combine an early-stopped DDPM wth another generative model such as VAE or GAN to obtain a new generative model. The combined model can outperform the original DDPM in terms of sample quality, sampling speed and training speed.
We can further accelerate the combined model by coupling it with other DDPM acceleration methods.
The combined model also has more control over the generation process. 
We demonstrate that we can control both high-level semantics and local details of the generated image by combining DDPMs with GANs.

Our method has achieved great acceleration effect and sample quality on single category datasets. 
However, for multi-category datasets, it still need many denoising steps to outperform the full-length DDPM in terms of sample quality due to the poor mode coverage of GANs.
In the future, we could combine ES-DDPM with other generative models that have better mode coverage to achieve high sample quality with less denoising steps on even complex multi-category datasets.
In addition, as a generative model, more studies are needed on how to prevent the model from learning biases in the training dataset and from misuse such as deception.

% \begin{ack}
% Wait to be written
% \end{ack}

\bibliographystyle{main}
\bibliography{main}

\appendix

\section{Appendix}

\subsection{ELBO Proof}
\label{sec:elbo_proof}
Prove 
\begin{align}
    \log p(\vx^{0}) \geq - (L_{\text{VAE}} + L_{\text{DDPM}})
\end{align}
in Equation~\ref{eqn:ddpm_elbo} in the main text.

\begin{proof}
Through variational inference, we have
\begin{align}
    \log p(\vx^0) &= \log \int p(\vx^{0:T'}, \vz) d\vx^{1:T'} d\vz \\
    &= \log \int q(\vx^{1:T'}, \vz | \vx^0) \frac{p(\vx^{0:T'}, \vz)}{q(\vx^{1:T'}, \vz | \vx^0)} d\vx^{1:T'} d\vz \\
    &\geq \int q(\vx^{1:T'}, \vz | \vx^0) \log \frac{p(\vx^{0:T'}, \vz)}{q(\vx^{1:T'}, \vz | \vx^0)} d\vx^{1:T'} d\vz \\
    &= \E_{q(\vx^{1:T'}, \vz | \vx^0)} \log \frac{p(\vx^{0:T'}, \vz)}{q(\vx^{1:T'}, \vz | \vx^0)}.
\end{align}
This is because $\log$ is a concave function and 
\begin{align}
    \int q(\vx^{1:T'}, \vz | \vx^0)  d\vx^{1:T'} d\vz = 1.
\end{align}

We plug in the definition that
\begin{align}
    p(\vx^{0:T'}, \vz) &= p(\vz) p_{\phi}(\vx^{T'} | \vz) p_{\theta}(\vx^{0:(T'-1)}|\vx^{T'}), \\
    q(\vx^{1:T'}, \vz | \vx^0) &= q(\vx^{1:T'} | \vx^0) q_{\psi}(\vz | \vx^0)
\end{align}
and obtain

\begin{align}
    \log p(\vx^{0}) 
    &\geq \E_{q(\vx^{1:T'}, \vz | \vx^0)} \log \frac{ p(\vz) p_{\phi}(\vx^{T'} | \vz) p_{\theta}(\vx^{0:(T'-1)}|\vx^{T'}) }{ q(\vx^{1:T'} | \vx^0) q_{\psi}(\vz | \vx^0) }\\
    &= \E_{q(\vx^{1:T'}, \vz | \vx^0)} [\log \frac{ p(\vz) }{ q_{\psi}(\vz | \vx^0) } + \log \frac{p_{\phi}(\vx^{T'} | \vz) p_{\theta}(\vx^{0:(T'-1)}|\vx^{T'}) }{q(\vx^{1:T'} | \vx^0) }]. 
\end{align}

We set the loss function as
\begin{align}
    L = - \E_{q(\vx^{1:T'}, \vz | \vx^0)} [\log \frac{ p(\vz) }{ q_{\psi}(\vz | \vx^0) } + \log \frac{p_{\phi}(\vx^{T'} | \vz) p_{\theta}(\vx^{0:(T'-1)}|\vx^{T'}) }{q(\vx^{1:T'} | \vx^0) }].
\end{align}

The first term is:
\begin{align}
    L_{1} &= -\E_{q(\vx^{1:T'}, \vz | \vx^0)} \log \frac{ p(\vz) }{ q_{\psi}(\vz | \vx^0) } \\
    &= \E_{q(\vx^{1:T'}, \vz | \vx^0)} \log \frac{ q_{\psi}(\vz | \vx^0) }{ p(\vz) }\\
    &= \E_{q(\vx^{1:T'} | \vx^0) q_{\psi}(\vz | \vx^0)} \log \frac{ q_{\psi}(\vz | \vx^0) }{ p(\vz) }\\
    &= D_{\text{KL}}(q_{\psi}(\vz | \vx^0)||p(\vz)).
\end{align}

The second term is:
\begin{align}
    L_{2} &= -\E_{q(\vx^{1:{T'}}, \vz | \vx_0)} \log \frac{p_{\phi}(\vx^{{T'}} | \vz) p_{\theta}(\vx^{0:({T'}-1)}|\vx^{{T'}}) }{q(\vx^{1:{T'}} | \vx_0) } \\
    &= \E_{q(\vx^{1:{T'}}, \vz | \vx_0)} \log \frac{q(\vx^{1:{T'}} | \vx_0) }{p_{\phi}(\vx^{{T'}} | \vz) p_{\theta}(\vx^{0:({T'}-1)}|\vx^{{T'}})} \\
    &= \E_{q(\vx^{1:{T'}}, \vz | \vx_0)} [\log \frac{ \prod_{t=1}^{T'} q(\vx^t|\vx^{t-1}) }{\prod_{t=1}^{T'} p_{\theta}(\vx^{t-1}|\vx^t)} - \log p_{\phi}(\vx^{{T'}} | \vz)] \\
    &= \E_{q(\vx^{1:{T'}}, \vz | \vx_0)} [\sum_{t=1}^{T'} \log \frac{q(\vx^t|\vx^{t-1})}{p_{\theta}(\vx^{t-1}|\vx^t)} - \log p_{\phi}(\vx^{{T'}} | \vz)] \\
    &= \E_{q(\vx^{1:{T'}}, \vz | \vx_0)} [\sum_{t=2}^{T'} \log \frac{q(\vx^t|\vx^{t-1})}{p_{\theta}(\vx^{t-1}|\vx^t)} + \log \frac{q(\vx^1|\vx^{0})}{p_{\theta}(\vx^{0}|\vx^1)} - \log p_{\phi}(\vx^{{T'}} | \vz)] \\
    &= \E_{q(\vx^{1:{T'}}, \vz | \vx_0)} [
    \sum_{t=2}^{T'} \log ( 
    \frac{ q(\vx^{t-1}|\vx^{t}, \vx^0) }{ p_{\theta}(\vx^{t-1}|\vx^t) } 
    \frac{ q(\vx^{t}|\vx^0) }{ q(\vx^{t-1}|\vx^0) } )
    + \log \frac{q(\vx^1|\vx^{0})}{p_{\theta}(\vx^{0}|\vx^1)} - \log p_{\phi}(\vx^{{T'}} | \vz)] \label{eqn:magic}\\
    &= \E_{q(\vx^{1:{T'}}, \vz | \vx_0)} [
    \sum_{t=2}^{T'} \log 
    \frac{ q(\vx^{t-1}|\vx^{t}, \vx^0) }{ p_{\theta}(\vx^{t-1}|\vx^t) } +
    \sum_{t=2}^{T'} \log \frac{ q(\vx^{t}|\vx^0) }{ q(\vx^{t-1}|\vx^0) } 
    + \log \frac{q(\vx^1|\vx^{0})}{p_{\theta}(\vx^{0}|\vx^1)} - \log p_{\phi}(\vx^{{T'}} | \vz)] \\
    &= \E_{q(\vx^{1:{T'}}, \vz | \vx_0)} [
    \sum_{t=2}^{T'} \log 
    \frac{ q(\vx^{t-1}|\vx^{t}, \vx^0) }{ p_{\theta}(\vx^{t-1}|\vx^t) } +
    \log \frac{ q(\vx^{{T'}}|\vx^0) }{ q(\vx^{1}|\vx^0) } 
    + \log \frac{q(\vx^1|\vx^{0})}{p_{\theta}(\vx^{0}|\vx^1)} - \log p_{\phi}(\vx^{{T'}} | \vz)] \\
    &= \E_{q(\vx^{1:{T'}}, \vz | \vx_0)} [
    \sum_{t=2}^{T'} \log 
    \frac{ q(\vx^{t-1}|\vx^{t}, \vx^0) }{ p_{\theta}(\vx^{t-1}|\vx^t) } +
    \log q(\vx^{{T'}}|\vx^0)
    - \log p_{\theta}(\vx^{0}|\vx^1) - \log p_{\phi}(\vx^{{T'}} | \vz)] \\
    &= \E_{q(\vx^{1:{T'}}, \vz | \vx_0)} [\log \frac{q(\vx^t|\vx_0)}{p_{\phi}(\vx^{{T'}} | \vz)} + 
    \sum_{t=2}^{T'} \log 
    \frac{ q(\vx^{t-1}|\vx^{t}, \vx^0) }{ p_{\theta}(\vx^{t-1}|\vx^t) } 
    - \log p_{\theta}(\vx^{0}|\vx^1)] \\
    &= \E_{q(\vx^{1:{T'}}, \vz | \vx_0)} \log \frac{q(\vx^t|\vx_0)}{p_{\phi}(\vx^{{T'}} | \vz)} + L_{\text{DDPM}},
    % D_{KL} (q(\vx^t|\vx_0) || p_{\phi}(\vx^{{T'}} | \vz))
\end{align}

where Line~\ref{eqn:magic} is because
\begin{align}
    &q(\vx^t|\vx^{t-1}) q(\vx^{t-1}|\vx^0) = q(\vx^{t-1}, \vx^{t}|\vx^0) = q(\vx^{t-1}|\vx^{t}, \vx^0) q(\vx^{t}|\vx^0) \Longrightarrow \\
    &q(\vx^t|\vx^{t-1}) = \frac{q(\vx^{t-1}|\vx^{t}, \vx^0) q(\vx^{t}|\vx^0)}{q(\vx^{t-1}|\vx^0)},
\end{align}
and
\begin{align}
    L_{\text{DDPM}} 
    &= \E_{q(\vx^{1:{T'}}, \vz | \vx_0)} [
    \sum_{t=2}^{T'} \log 
    \frac{ q(\vx^{t-1}|\vx^{t}, \vx^0) }{ p_{\theta}(\vx^{t-1}|\vx^t) } 
    - \log p_{\theta}(\vx^{0}|\vx^1)] \\
    &= \sum_{t=2}^{T'} \E_{q(\vx^{1:{T'}}, \vz | \vx_0)} \log 
    \frac{ q(\vx^{t-1}|\vx^{t}, \vx^0) }{ p_{\theta}(\vx^{t-1}|\vx^t) } 
    - \E_{q(\vx^{1:{T'}}, \vz | \vx_0)} \log p_{\theta}(\vx^{0}|\vx^1) \\
    &= \sum_{t=2}^{T'} 
    \E_{ q(\vx^t|\vx^0) q(\vx^{t-1}|\vx^0, \vx^t) q(\vx^{1:{T'}} \setminus \{\vx^{t-1}, \vx^{t}\}, \vz|\vx^0, \vx^{t-1}, \vx^t)} 
    \log \frac{ q(\vx^{t-1}|\vx^{t}, \vx^0) }{ p_{\theta}(\vx^{t-1}|\vx^t) } \\
        & \;\;\;\;- \E_{q(\vx^{2:{T'}}, \vz | \vx^0, \vx^1)q(\vx^{1} | \vx^0)} \log p_{\theta}(\vx^{0}|\vx^1) \\
    &= \sum_{t=2}^{T'} 
    \E_{ q(\vx^t|\vx^0) } 
    D_{\text{KL}}( q(\vx^{t-1}|\vx^{t}, \vx^0) || p_{\theta}(\vx^{t-1}|\vx^t) ) - \E_{q(\vx^{1} | \vx^0)} \log p_{\theta}(\vx^{0}|\vx^1).
\end{align}

The first term in $L_2$ is
\begin{align}
    \E_{q(\vx^{1:T'}, \vz | \vx^0)} \log \frac{q(\vx^{T'}|\vx^0)}{p_{\phi}(\vx^{T'} | \vz)} &= \E_{q(\vx^{1:(T'-1)}, \vz | \vx^0, \vx^{T'}) q(\vx^{T'} | \vx^0)} \log \frac{q(\vx^{T'}|\vx^0)}{p_{\phi}(\vx^{T'} | \vz)} \\
    &= \E_{q(\vx^{1:(T'-1)}, \vz | \vx^0, \vx^{T'})} D_{\text{KL}} (q(\vx^{T'}|\vx^0) || p_{\phi}(\vx^{T'} | \vz)) \\
    &= \E_{q(\vx^{1:(T'-1)} | \vx^0, \vx^{T'}, \vz) q(\vz | \vx^0, \vx^{T'})} D_{\text{KL}} (q(\vx^{T'}|\vx^0) || p_{\phi}(\vx^{T'} | \vz)) \\
    &= \E_{q_{\psi}(\vz | \vx^0)} D_{\text{KL}} (q(\vx^{T'}|\vx^0) || p_{\phi}(\vx^{T'} | \vz)).
\end{align}

Now the total loss is
\begin{align}
    L &= L_1 + L_2 \\
    &= D_{\text{KL}}(q_{\psi}(\vz | \vx^0)||p(\vz)) + \E_{q_{\psi}(\vz | \vx^0)} D_{\text{KL}} (q(\vx^{T'}|\vx^0) || p_{\phi}(\vx^{T'} | \vz)) + L_{\text{DDPM}} \\
    &= L_{\text{VAE}} + L_{\text{DDPM}},
\end{align}
where $L_{\text{VAE}}$ is defined as
\begin{align}
    L_{\text{VAE}} = D_{\text{KL}}(q_{\psi}(\vz | \vx^0)||p(\vz)) + \E_{q_{\psi}(\vz | \vx^0)} D_{\text{KL}} (q(\vx^{T'}|\vx^0) || p_{\phi}(\vx^{T'} | \vz)).
\end{align}

We have finished the proof that
\begin{align}
    \log p(\vx^{0}) \geq -L = - (L_{\text{VAE}} + L_{\text{DDPM}}).
\end{align}

\end{proof}

\subsection{Conditional Generation}
\label{sec:conditional_generation}
In the case of conditional generation, we use $\vc$ to denote the condition. It contains some information about the clean image $\vx^0$. For example, $\vc$ could be the class of $\vx^0$, a low resolution image of $\vx^0$, or a grayscale image of $\vx^0$.

We adapt the encoding process in Equation~\ref{eqn:encoding_process} to
\begin{align}
\label{eqn:conditional_encoding_process}
    q(\vx^{1:T'}, \vz | \vx^0, \vc) = q(\vx^{1:T'} | \vx^0, \vc) q_{\psi}(\vz | \vx^{1:T'}, \vx^0, \vc) = q(\vx^{1:T'} | \vx^0) q_{\psi}(\vz | \vc),
\end{align}
where $q_{\psi}(\vz | \vc)$ is the encoder. We set it to be a Gaussian distribution whose mean and standard deviation are parameterized by a neural network. 
$q(\vx^{1:T'} | \vx^0)$ is the same as Equation~\ref{eqn:partial_diffusion_process}. We then adapt the sampling process in Equation~\ref{eqn:generation_process} to
\begin{align}
\label{eqn:conditional_generation_process}
    p(\vx^{0:T'}, \vz | \vc) = p(\vz) p_{\phi}(\vx^{T'} | \vz, \vc) p_{\theta}(\vx^{0:(T'-1)}|\vx^{T'}, \vc),
\end{align}
where $p(\vz)$ is assumed to follow standard Gaussian distribution, $p_{\phi}(\vx^{T'} | \vz, \vc)$ is the decoder parameterized by a neural network, and $p_{\theta}(\vx^{0:(T'-1)}|\vx^{T'}, \vc)$ is the reverse process of the partial diffusion process in Equation~\ref{eqn:partial_diffusion_process}. It is defined as
\begin{align}
\label{eqn:conditional_partial_reverse_process}
    p_{{\theta}}(\vx^{0:(T'-1)}|\vx^{T'}, \vc)=\prod_{t=1}^{T'} p_{{\theta}}(\vx^{t-1}|\vx^t, \vc),
    \text{ where }
    p_{{\theta}}(\vx^{t-1}|\vx^t, \vc) = \gN(\vx^{t-1};\bm{\mu}_{{\theta}}(\vx^t, \vc, t), \sigma_t^2\mI).
\end{align}

Through variational inference, we can prove that
\begin{align}
    \log p(\vx^{0}|\vc) \geq - (L_{\text{VAE}} + L_{\text{DDPM}}), 
\end{align}
\begin{align}
\text{where }
    L_{\text{VAE}} &= D_{\text{KL}}(q_{\psi}(\vz | \vc)||p(\vz)) + \E_{q_{\psi}(\vz | \vx^0)} D_{\text{KL}} (q(\vx^{T'}|\vx^0) || p_{\phi}(\vx^{T'} | \vz, \vc)), \\
    L_{\text{DDPM}} &= \sum_{t=2}^{T'} 
    \E_{ q(\vx^t|\vx^0) } 
    D_{\text{KL}}( q(\vx^{t-1}|\vx^{t}, \vx^0) || p_{\theta}(\vx^{t-1}|\vx^t, \vc) ) 
    - \E_{q(\vx^{1} | \vx^0)} \log p_{\theta}(\vx^{0}|\vx^1, \vc).
\end{align}
The proof is the same as Section~\ref{sec:elbo_proof}.
Now we have two loss terms, $L_{\text{DDPM}}$ and $L_{\text{VAE}}$. This means that we can train a conditional DDPM and a VAE-like model separately, and then combine them together to obtain a conditional generative model. 
First, The loss term $L_{\text{DDPM}}$ is the same of a normal conditional DDPM, except that we do not need to train the full chain range from $1$ to $T$. Instead, we only need to train the first several steps range from $1$ to $T'$. 

Next, we will show that the loss term $L_{\text{VAE}}$ amounts to train a VAE-like conditional generative model.
Similarly to Equation~\ref{eqn:sample_xT}, we can set $p_{\phi}(\vx^{T'} | \vz, \vc)$ to
\begin{align}
\label{eqn:conditional_sample_xT}
    p_{\phi}(\vx^{T'} | \vz, \vc) \sim N(\vx^{T'}; \sqrt{\bar{\alpha}_t} \vf_{\phi} (\vz, \vc), (1-\bar{\alpha}_t)\mI).
\end{align}
Then 
\begin{align}
    D_{\text{KL}} (q(\vx^{T'}|\vx^0) || p_{\phi}(\vx^{T'} | \vz, \vc)) = C_1 ||\vx^0-\vf_{\phi} (\vz, \vc)||^2 + C_2,
\end{align}
where $C_1, C_2$ are some scalar constants.
Now the second term in $L_{\text{VAE}}$ means that the decoder $\vf_{\phi} (\vz, \vc)$ need to reconstruct the original image $\vx_0$, based on the latent code $\vz$ sampled from the encoder $q_{\psi}(\vz | \vc)$ and the condition $\vc$. 
Therefore, $L_{\text{VAE}}$ can be optimized by following a similar procedure to train a VAE on clean images $\vx^0$, except that we need to change the input of the encoder $q_{\psi}$ from $\vx^0$ to  $\vc$, and allow the decoder $\vf_{\phi}$ to utilize the condition $\vc$ when reconstructing the original image $\vx^0$.

\paragraph{Sampling Procedure.} After training the conditional DDPM and the VAE-like model separately, we can combine them to form a conditional generative model. We can use the following equation to generate samples given condition $\vc$:
\begin{align}
\label{eqn:real_conditional_generation_process}
    p(\vx^{0:T'}, \vz | \vc) = q_{\psi}(\vz | \vc) p_{\phi}(\vx^{T'} | \vz, \vc) p_{\theta}(\vx^{0:(T'-1)}|\vx^{T'}, \vc).
\end{align}
First sample the latent code $\vz$ from 
$q_{\psi}(\vz | \vc)$
% the standard Gaussian distribution
, then use the decoder of the VAE to generate an image $\vf_{\phi} (\vz, \vc)$. Next, sample $\vx^{T'}$ from the distribution described in Equation~\ref{eqn:conditional_sample_xT}. Finally, use the DDPM to sample $\vx^{t-1}$ from $p_{\bm{\theta}}(\vx^{t-1}|\vx^t, \vc)$ for $t=T',T'-1,\cdots,1$, and output the generated image $\vx^0$.

% Note that the first two terms in Equation~\ref{eqn:real_conditional_generation_process}, $q_{\psi}(\vz | \vc) p_{\phi}(\vx^{T'} | \vz, \vc)$, is equivalent to a model that can generate an clean image given the condition $\vc$: $\vg(\vc)$. To sample $\vx^{T'}$, we only need to replace $\vf_{\phi} (\vz, \vc)$ in Equation~\ref{eqn:conditional_sample_xT} with $\vg(\vc)$.
During sampling, we do not need to restrict how $\vf_{\phi} (\vz, \vc)$ is trained, or what kind of model it is.
It could be a conditional GAN or any other conditional generation model.

\subsection{Computational Cost}
\label{sec:computational_cost}
On the CelebA dataset, we train the StyleGAN2, the ES-DDPM and the full-length DDPM on 8 NVIDIA RTX 2080Ti GPUs. 
For CelebA-64, the StyleGAN2 and the full-length DDPM are trained for $10^5$ (around 25 hours) iterations and $9 \times 10^5$ (around 80 hours) respectively. 
For CelebA-128, the StyleGAN2 and the full-length DDPM are trained for $2.4 \times 10^5$ (around 183 hours) iterations and $4 \times 10^6$ (around 448 hours) respectively. 
As for ES-DDPMs, their training time of each iteration is the same as full-length DDPMs and can be inferred from Figure~\ref{fig:celeba_accelerate_training}. 
For image generation experiments, it takes around one day for the 1000-step DDPM to generate $50000$ random Celeba-128 images on 8 NVIDIA RTX 2080Ti GPUs. Other image generation experiment on CelebA takes less time.

For the ImageNet-64 dataset, we use pre-trained DDPM models and pre-generated BigGAN-deep images from the work~\citep{dhariwal2021diffusion}.
It takes about 8 hours for the 250-step DDPM to generate $50000$ random images on 4 NVIDIA A100 GPUs. Generation time of different length ES-DDPMs can be inferred accordingly.

For LSUN-Bedroom and LSUN-Cat datasets, we use pre-trained DDPM models and pre-generated StyleGAN images from the work~\citep{dhariwal2021diffusion}.
It takes about one day for the 1000-step DDPM to generate $50000$ random LSUN-Bedroom or LSUN-Cat images on 24 NVIDIA A100 GPUs. Generation time of different length ES-DDPMs can be inferred accordingly.

\subsection{DDPM Acceleration Experiement Results on LSUN Datasets}
\label{sec:lsun_ddpm_acceleration}
\begin{table}[thb]
% \parbox{0.49\columnwidth}{ 
\centering
\caption{FID comparison of different DDPM acceleration methods on the LSUN-Bedroom dataset.}
\label{tbl:lsun_bedroom_acceleration}
% \resizebox{0.49\columnwidth}{!}{
% \small{
\begin{tabular}{c|c|c|c|c|c|c}
    \Xhline{1pt}
	Denoising Steps       & 500 & 100 & 50 & 20 & 10 & 5 \\
	\hline \hline
    ES-DDPM($T'=100$)+StyleGAN+TR(\textbf{Ours})      && \textbf{1.85} & \textbf{1.83} & \textbf{2.03} & \textbf{2.67} & \textbf{3.84} \\
% 	\cline{1-10}
	DDIM~\citep{song2020denoising} && 6.62 & 6.75 & 8.89 & 16.95 &  \\
% 	\cline{1-10}
	PNDM~\citep{liu2022pseudo}  && 6.91 & 6.44 & 5.68 & 6.99 & 12.6 \\ 
	FastDPM~\citep{kong2021fast} && 7.98 & 8.37 & 9.86 & 19.07 &  \\
	TDPM-GAN~\citep{zheng2022truncated} &3.95&&4.10&&& \\
	TDPM-CT~\citep{zheng2022truncated} &4.01&&4.95&&& \\
% 	\cline{1-10}
	\Xhline{1pt}
\end{tabular}
% }
% }
% }
\end{table}

\begin{table}[thb]
% \parbox{0.49\columnwidth}{ 
\centering
\caption{FID comparison of different DDPM acceleration methods on the LSUN-Cat dataset. ``*'' means the images for evaluation are generated by ourselves for this method.}
\label{tbl:lsun_cat_acceleration}
% \resizebox{0.49\columnwidth}{!}{
% \small{
\begin{tabular}{c|c|c|c|c|c}
    \Xhline{1pt}
	Denoising Steps       & 100 & 50 & 20 & 10 & 5 \\
	\hline \hline
    ES-DDPM($T'=100$)+StyleGAN2+TR(\textbf{Ours}) & \textbf{5.47} & \textbf{5.88} & \textbf{7.27} & \textbf{9.72} & \textbf{13.48} \\
% 	\cline{1-10}
	DDIM*~\citep{song2020denoising} & 7.43 & 8.80 & 12.40 & 20.11 & 48.39 \\
% 	\cline{1-10}
	TR*~\citep{nichol2021improved}  & 9.88 & 14.60 & 28.86 & 56.62 & 114.00 \\ 
% 	\cline{1-10}
	\Xhline{1pt}
\end{tabular}
% }
% }
% }
\end{table}

\subsection{Additional Controllable Generation Experiments}
\label{sec:additional_controllable_generation_exps}
\begin{figure}[thb]
    % \centering
    % \vspace{-3em}
    % \label{fig:mlp}
    \includegraphics[width=1\textwidth]{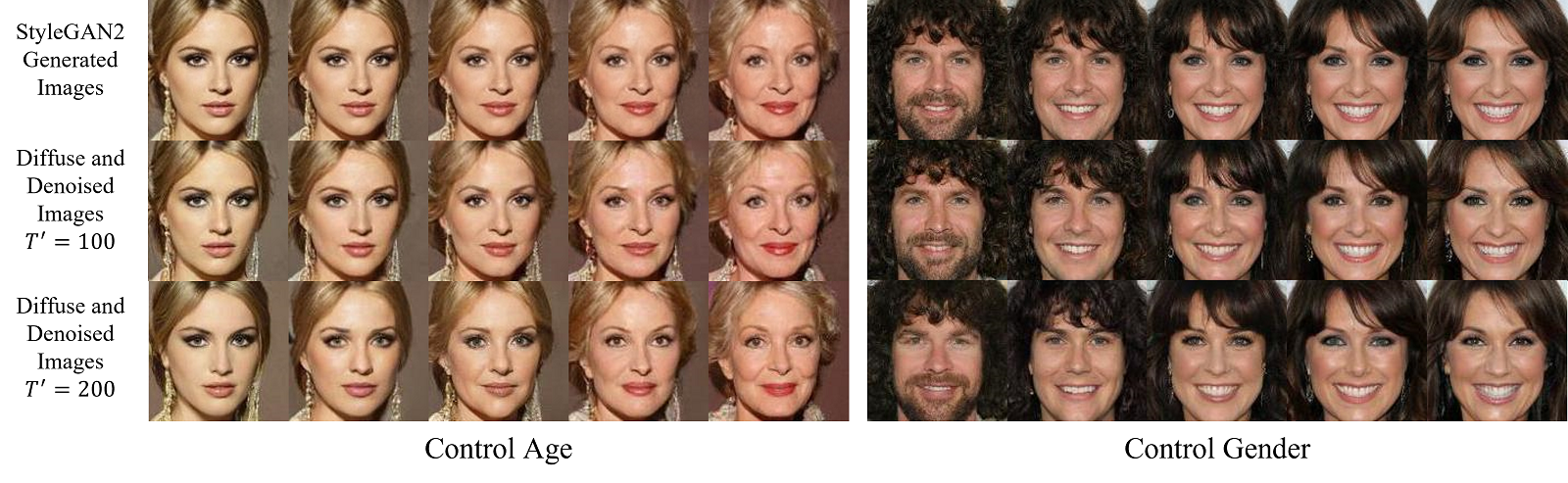}
    \caption{On CelebA-128, ES-DDPM can preserve high-level semantics in the images generated by StyleGAN2 for not too large $T'$'s. Hence our combined model can control high-level semantics such as gender and age in generated images.}
    \label{fig:face_editing}
    % \vspace{-1em}
\end{figure}

\begin{figure}[thb]
    % \centering
    % \vspace{-3em}
    % \label{fig:mlp}
    \includegraphics[width=0.99\textwidth]{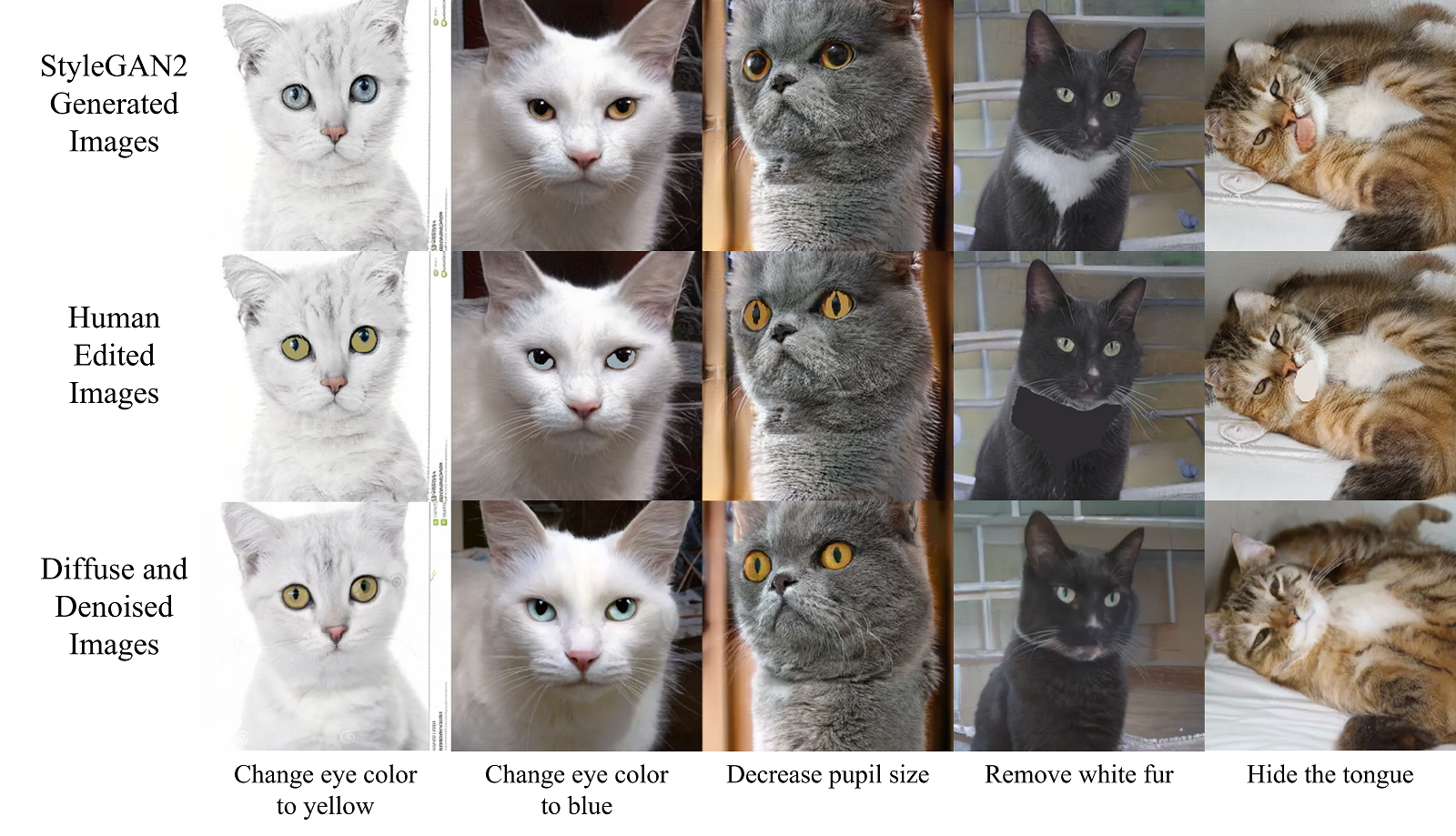}
    \caption{On LSUN-Cat-256, we perform stroke-based local editing on StyleGAN2 generated cats. ES-DDPM($T'=300$) can refine the editing we made and remove artifacts.}
    \label{fig:cat_editing}
    % \vspace{-1em}
\end{figure}
Figure~\ref{fig:face_editing} shows that we can use StyleGAN2 to control the age and the gender of the generated face, and the ES-DDPM can preserve these high-level semantic information.
In this way, we can control high-level semantics for our combined model.
Other works~\citep{preechakul2021diffusion, pandey2022diffusevae} have also shown that it is possible to combine DDPMs and VAEs to control high-level semantics of the generated images. 
% but unlike our method, their methods have little or no acceleration effect on the sampling process of the DDPM.
Compared with them, our method can achieve better sample quality using the same number of denoising steps as shown in Table~\ref{tbl:celeba_64} and Table~\ref{tbl:celeba_acceleration}.
In addition, these methods need to alter the training process of the DDPM and train a DDPM that is specific to the other model they want to combine DDPM with.
On the contrary, our method can combine a pre-trained DDPM with any other generative model that has the ability to control high-level semantics to achieve high-level controllable generation for the DDPM.

Figure~\ref{fig:cat_editing} shows that we can perform stroke-based local editing on the cat images generated by StyleGAN2, and then use the ES-DDPM to refine it, mitigate artifacts and make the images photo-realistic.
This local editing technique is the same as the one proposed in the work~\citep{meng2021sdedit}.

\end{document}